\documentclass[letterpaper]{article} 
\usepackage{aaai2026}  
\usepackage{times}  
\usepackage{helvet}  
\usepackage{courier}  
\usepackage[hyphens]{url}  
\usepackage{graphicx} 
\usepackage{natbib}  
\usepackage{caption} 
\usepackage{algorithm}
\usepackage{algorithmic}
\usepackage{xspace}
\usepackage{amssymb}
\usepackage{amsmath}
\usepackage{subcaption}
\usepackage{multirow}
\usepackage{booktabs}
\usepackage{adjustbox}
\usepackage{mathrsfs}
\newcommand{\method}{\texttt{SAMCL}\xspace}

\usepackage{newfloat}
\usepackage{listings}
\DeclareCaptionStyle{ruled}{labelfont=normalfont,labelsep=colon,strut=off} 
\floatstyle{ruled}
\newfloat{listing}{tb}{lst}{}
\floatname{listing}{Listing}
\title{SAMCL: Empowering SAM to Continually Learn from Dynamic Domains with Extreme Storage Efficiency}
\author{    
    Zeqing Wang\textsuperscript{\rm 1,\rm 2}, \, Kangye Ji\textsuperscript{\rm 1,\rm3}, \, Di Wang\textsuperscript{\rm 1}, \, Haibin Zhang\textsuperscript{\rm 4}, \, Fei Cheng\textsuperscript{\rm 1}\thanks{Corresponding Author.}
}
\affiliations{
    \textsuperscript{\rm 1}School of Computer Science and Technology, \, Xidian University \\ \textsuperscript{\rm 2}College of Design and Engineering, \, National University of Singapore \\
    \textsuperscript{\rm 3}Tsinghua Shenzhen International Graduate School, \, Tsinghua University \\
    \textsuperscript{\rm 4}School of Cyber Engineering, \, Xidian University  \\
     zeqing.wang@u.nus.edu, \, jky25@mails.tsinghua.edu.cn, \, wangdi@xidian.edu.cn, \, \\ hbzhang@mail.xidian.edu.cn, \, chengfei@xidian.edu.cn
}

\usepackage{bibentry}

\begin{document}

\maketitle

\begin{abstract}
Segment Anything Model (SAM) struggles in open-world scenarios with diverse domains.
In such settings, naive fine-tuning with a well-designed learning module is inadequate and often causes \textit{catastrophic forgetting} issue when learning incrementally. 
To address this issue, we propose a novel continual learning (CL) method for SAM, termed \method. Rather than relying on a fixed learning module, our method decomposes incremental knowledge into separate modules and trains a selector to choose the appropriate one during inference.
However, this intuitive design introduces two key challenges: ensuring effective module learning and selection, and managing storage as tasks accumulate.
To tackle these, we introduce two components: \textsl{AugModule} and \textsl{Module Selector}.
\textsl{AugModule} reduces the storage of the popular LoRA learning module by sharing parameters across layers while maintaining accuracy. It also employs heatmaps—generated from point prompts—to further enhance domain adaptation with minimal additional cost.
\textsl{Module Selector} leverages the observation that SAM’s embeddings can effectively distinguish domains, enabling high selection accuracy by training on low-consumed embeddings instead of raw images.
Experiments show that \method outperforms state-of-the-art methods, achieving only $0.19\%$ forgetting and at least $2.5\%$ gain on unseen domains. Each \textsl{AugModule} requires just $0.233$ MB, reducing storage by at least $24.3\%$ over other fine-tuning approaches. The buffer storage for \textsl{Module Selector} is further reduced by up to $256\times$.
\end{abstract}

\begin{links}
    \link{Code}{https://github.com/INV-WZQ/SAMCL}
    \link{Extended version}{https://arxiv.org/abs/2412.05012}
\end{links}

\section{Introduction}
Segment Anything Model (SAM) has achieved notable success in image segmentation~\cite{sam}. This success stems from its extensive pre-training on a closed-set dataset and its distinctive architecture that enables effective integration of image and prompt information. However, when deployed in open-world environments~\cite{notperfect, Tang_2025_CVPR}, conventional SAM exhibits significant limitations in segmenting instances across diverse domains, including camouflaged, shadowed, and medical domains, thereby restricting its practical applicability.

Recent studies have identified this flaw and utilized fine-tuning strategies using carefully designed learning modules to enhance SAM’s performance~\cite{SAM-Adapter, AutoSAM}. However, these modules remain fixed throughout the entire learning process, meaning that all additional knowledge is stored in predetermined locations. This design is inherently limited in real scenarios where the model encounters incremental data and needs continual learning (CL) manner~\cite{CLSAM}. In such scenarios, fine-tuning on new data often results in significant forgetting of previous knowledge, leading to the issue known as \textit{catastrophic forgetting}~\cite{forgetting}. 

To address this issue, we propose a novel CL method for SAM, termed \method. At a high level, \method allocates different knowledge into distinct modules for learning, and trains an accurate selector that dynamically chooses the most appropriate module during inference. This design mitigates inter-task interference and effectively reduces forgetting.

While conceptually straightforward~\cite{PNN, O-Lora, Expert_gate}, this approach faces two main challenges in the CL setting for SAM: (1) ensuring effective knowledge learning within each module and maintaining high selection accuracy, and (2) managing the increasing storage overhead from accumulating modules and the rising cost of training an accurate selector as tasks grow.

To overcome these challenges, we introduce an effective and efficient learning module, \textsl{AugModule}, and selector, \textsl{Module Selector}, for \method. 

\textbf{Firstly}, \textsl{AugModule}—comprising \textsl{SLoRA} and \textsl{Prompt Augmentation}—efficiently adapts SAM to a new domain. Unlike vanilla LoRA~\cite{Lora} which utilizes independent matrices $A$ and $B$ for each layer, 
\textsl{SLoRA} \underline{s}hares a single matrix $A$ in \underline{LoRA}s across all layers without compromising accuracy. \textsl{Prompt Augmentation}, which transforms point-type prompts into heatmaps, augments prompt utilization without retraining the mask decoder in SAM. Compared with learning modules in other fine-tuning methods, \textsl{AugModule} achieves high accuracy across all tested domains, with parameter storage costs of only $0.233$ MB, resulting in at least $24.3$\% reduction. 

\textbf{Secondly}, we propose a lightweight \textsl{Module Selector} consisting of only four linear layers to select the appropriate module during inference. We exploit a novel observation that embeddings in the image encoder can accurately differentiate between domains. This enables us to store only a few low-consumed embeddings in the buffer, which are $256\times$ lower than storing raw images~\cite{CLSAM, DER, ER}, for training \textsl{Module Selector} while maintaining high accuracy in module selection. This approach not only mitigates the forgetting problem but also enhances knowledge transfer to unseen domains. 

Together, \method empowers SAM with CL ability to learn from incremental and dynamic domains. Experimental results show that \method limits average accuracy loss to just $0.19\%$—significantly lower than state-of-the-art methods—while achieving at least a $2.5\%$ improvement in accuracy when transferring to unseen environments.

In summary, our main contributions are as follows:
\begin{itemize}
    \item We design \textsl{AugModule} for SAM to effectively learn from any new domain with ultra-low storage cost.
    \item We store a small set of low-consumed embeddings in the buffer to train a lightweight \textsl{Module Selector}, leveraging the inherent ability of SAM to effectively distinguish between domains for module selection during inference.
    \item We design a new experimental setup that can comprehensively evaluate the performance of SAM during CL. Experiments show the excellent performance of \method in learning, maintaining, and transferring knowledge.
\end{itemize}

\section{Related Works}\label{sec:related}
\begin{figure*}[ht]
    \centering
    \includegraphics[width=1.\textwidth]{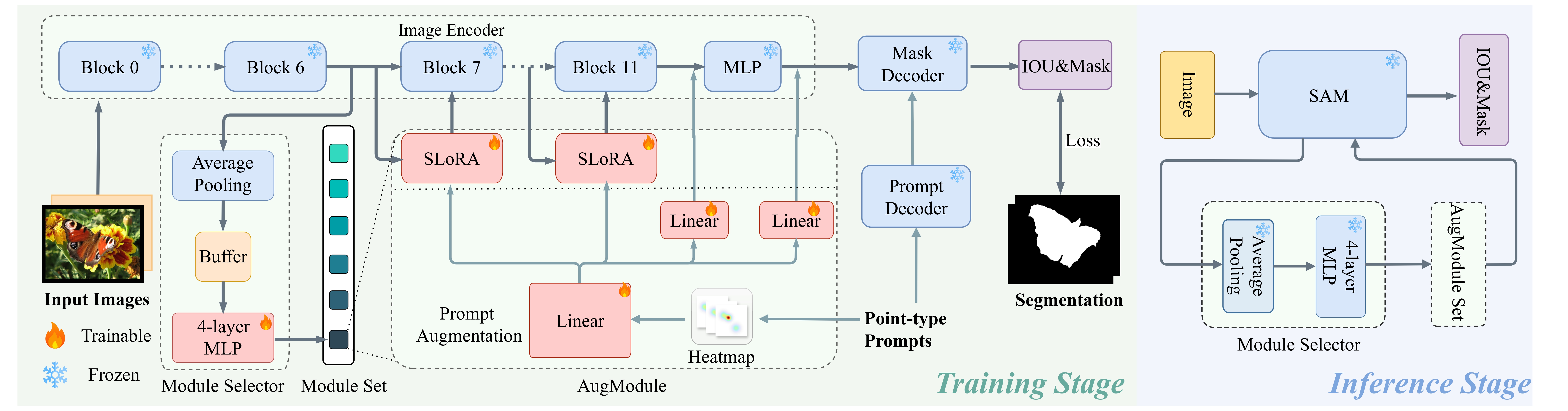}
    \caption{Overview of \method with SAM. During training, \method uses a new \textsl{AugModule} to learn from a new domain. All modules are stored in a module set. Meanwhile, \textsl{Module Selector} is trained on a few stored embeddings from the image encoder. During inference, \method extracts latent embeddings from the image encoder to select an appropriate module by \textsl{Module Selector} and continual inference by SAM with the selected module. A detailed illustration of the inference process is provided in the Appendix~B.1.}
    \label{fig:method}
\end{figure*}

\paragraph{Segment Anything Model.} 
SAM~\cite{sam} demonstrates strong performance in image segmentation using interactive prompts. SAM2~\cite{sam2} extends this capability to video. In this paper, we focus on image segmentation. Due to their architectural similarity and SAM’s popularity in fine-tuning, we adopt SAM as the base model to present our method. A detailed discussion of applying our method to SAM2 is included in the Appendix~B.4.

\paragraph{Fine-tuning SAM.} 
To address SAM’s limitations in challenging domains such as camouflaged, shadow, and medical domains, a common strategy is to freeze parts of SAM and add extra modules for fine-tuning. OCRT~\cite{Tang_2025_CVPR} extracts informative low-level and high-level image representations to facilitate adaptation to unseen domains. SAM-LST~\cite{SAM-LST} appends a ResNet~\cite{resnet} to image encoder and trains mask decoder. AutoSAM~\cite{AutoSAM} replaces the prompt encoder with a new network while freezing the rest of SAM. SAM-Adapter~\cite{SAM-Adapter} fine-tunes the mask decoder and uses task-specific knowledge to guide performance within the frozen image encoder. However, most of them overlook the need to handle dynamic environments rather than a single domain. Moreover, fine-tuning the mask decoder or adding large networks incurs significant storage overhead for each domain.

\paragraph{LoRA-type Fine-tuning.}
As a foundation model, the widely used LoRA technique~\cite{Lora} is applicable to SAM~\cite{samed}. LoRA employs two low-rank matrices, $A$ and $B$, to fine-tune transformer blocks. This process can be articulated as follows:
\begin{equation}
    Y_i = (W_i + B_iA_i)X_i,
  \label{eq:Lora}
\end{equation}
where $Y$, $X$, $W$, $A$ and $B$ represent output, input, pre-trained weight, and matrices $A$ and $B$ in LoRA. Note that $i$ represents $i$-th block in the pre-trained model. 

While LoRA achieves strong performance with few parameters, AsymmLoRA~\cite{asymmetry} further reduces redundancy by freezing matrix $A$ across all LoRAs, lowering trainable parameters. However, it still incurs storage cost for matrix $A$. HydraLoRA~\cite{HydraLoRA} reduces storage by sharing a single matrix $A$ across multiple LoRAs within a layer. In contrast, our method, \textit{SLoRA}, uses one LoRA for each layer and shares matrix $A$ across all layers, minimizing storage further. Although some similar works, such as ShareLoRA~\cite{sharelora}, employ a comparable paradigm in the field of natural language processing, we broaden the application of this paradigm and confirm its efficacy in visual segmentation models, including SAM and SAM2. More importantly, we show that \textsl{Prompt Augmentation} offers a simple yet effective way to enhance the performance of \textsl{SLoRA}, without additional training on the mask decoder in SAM.

\paragraph{Continual Learning.} 
CL methods are categorized into regularization, replay, and architecture-based approaches~\cite{2023survey}. Regularization-based methods~\cite{EWC, LWF, LightCL} preserve prior knowledge by constraining important historical parameters. Most continual segmentation approaches~\cite{SPPA, LAG, Tang_2025_ICCV, tang2024bootstrapsegmentationfoundationmodel} follow this paradigm, leveraging representation learning or knowledge distillation to reduce segmentation confusion~\cite{survey_seg}.
Replay-based methods~\cite{DER, ER} retain and replay past samples when training on new tasks. While effective, they incur substantial storage overhead due to the large buffer.
Both of them typically rely on a fixed network, leading to entanglement between old and new knowledge. 

To mitigate inter-task interference from shared parameters, architecture-based methods~\cite{PNN, Packnet, Ada-QPacknet, TinySubNets} allocate separate modules per task. 
Expert Gate~\cite{Expert_gate} uses a pre-trained gate to select experts for task-agnostic inference but introduces high overhead. O-LoRA~\cite{O-Lora} assigns tasks to orthogonal low-rank subspaces to reduce interference but fails to ensure strict orthogonality.

MoDA~\cite{CLSAM} uses extra tokens as selectors to choose the proper module for SAM. However, it requires storing raw samples to train selectors, incurring high storage costs. Additionally, when paired with encoder-finetuning methods, MoDA must run the image encoder twice during inference, increasing latency. In contrast, \method avoids both issues by training on lightweight embeddings instead of raw images and eliminating redundant forward passes.

\section{Preliminaries}\label{sec:definition}
\paragraph{Problem Formulation.}
In the CL setting, tasks $\{1, 2,.., T\}$ are executed sequentially with no access to previous data when training on a new task, except for replay data in replay-based methods~\cite{DER}. The \textsl{t}-th task is described as $\mathcal D_t=\{(x_t^i, y_t^i)\}^{n_t}_{i=1}$, where $n_t$ represents the number of samples for the \textsl{t}-th task, and $(x_t^i, y_t^i)$ denotes the input-label pair. For using SAM to segment an instance with the help of prompts, $x_t^i \in \mathbb{R}^{C\times H \times W}$ represents an image, and $y_t^i \in \{0, 1\}^{H \times W}$ is the corresponding mask based on the prompt. $\theta_{t}$ denotes the model after training on the \textsl{t}-th task. The objective of the CL is to obtain the model $\theta_T$ that performs well on all seen domains from $D_1$ to $D_T$, while also exhibiting strong generalization to unseen domains.

\paragraph{Evaluation Metrics.} To evaluate segmentation performance, we employ three common metrics: mean absolute error (mMAE), mean F1 score (mF1), and mean intersection over union (mIoU). We denote the accuracy of these metrics for the \textit{j}-th task, after training on the \textit{i}-th task, as $a_{i,j}$.
To evaluate the CL performance in terms of learning, maintaining, and transferring knowledge, we follow previous works~\cite{GEM, 2023survey} and define three key metrics as follows: (1) \textbf{Average Accuracy (AA)}, defined as AA $=\frac{1}{T}\sum_{i=1}^T a_{T,i}$, to evaluate the segmentation performance of all tasks after CL; (2) \textbf{Forgetting Measure (FM)}, defined as FM $=\frac{1}{T}\sum_{i=1}^{T} a_{i,i} - a_{T,i}$, to measure the forgetting degree after CL; (3) \textbf{Forward Transfer (FT)}, defined as FT $=\frac{1}{T-1}\sum_{i=1}^{T-1} a_{i, i+1}$, to evaluate the transferring ability to unseen domains during CL.

\section{Method}\label{sec:method}
\method allocates different knowledge into distinct modules for learning, and trains an accurate selector to dynamically choose the most appropriate module during inference. The overview is shown in Figure~\ref{fig:method}. To ensure both effectiveness and storage efficiency for the learning module and the selector, we introduce two key components in this section: \textsl{AugModule} (Section~\ref{sec:augmodule}) and \textsl{Module Selector} (Section~\ref{sec:task}). 

\subsection{AugModule}\label{sec:augmodule}
\begin{figure}[ht]
    \centering
    \includegraphics[width=0.85\linewidth]{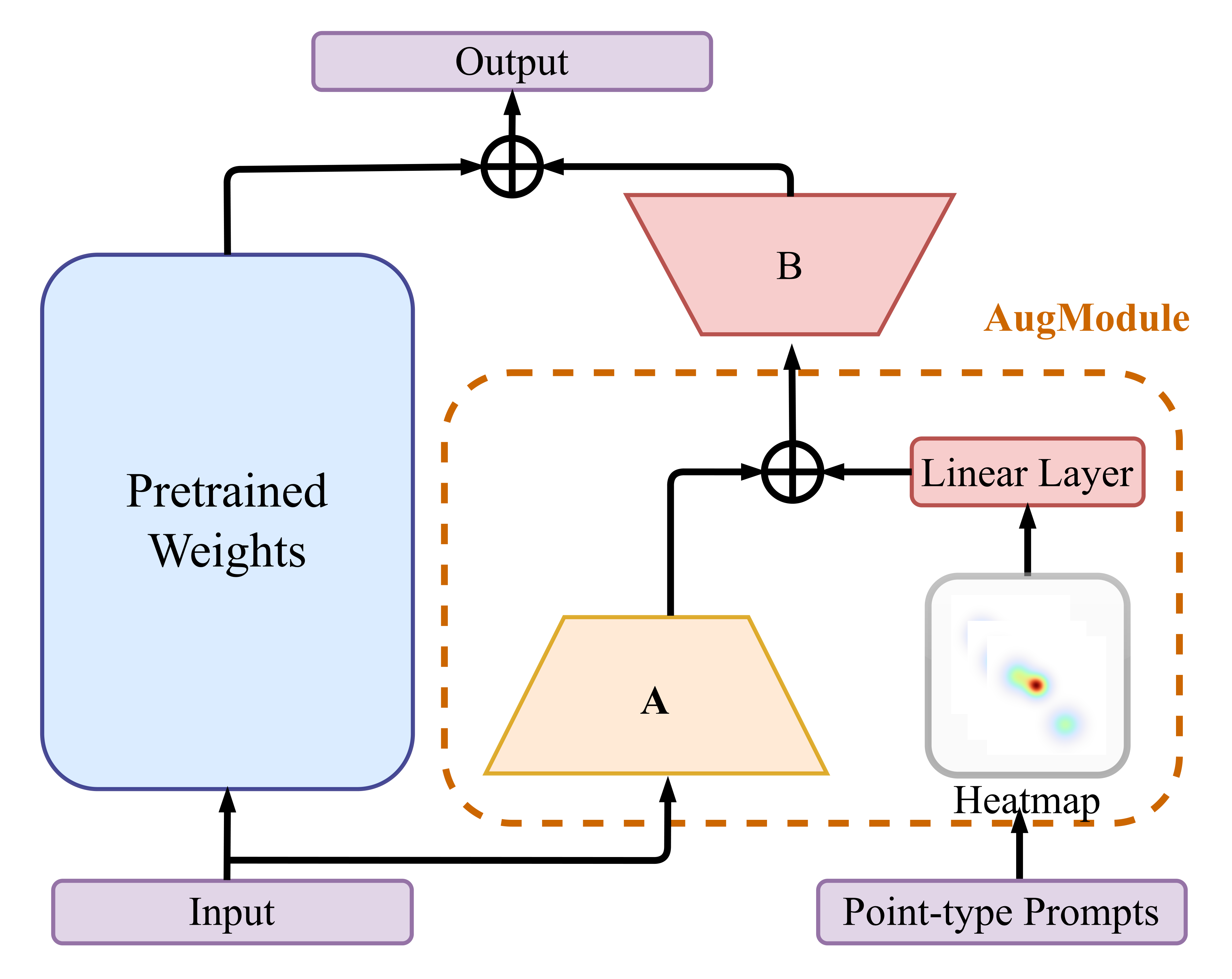}
    \caption{Illustration of \textsl{AugModule}, integrating \textsl{SLoRA} and \textsl{Prompt Augmentation}. \textsl{SLoRA} shares matrix $A$ across all LoRAs for efficient adaptation. \textsl{Prompt Augmentation} converts point prompts into heatmaps, injected via linear transformation for dimensional alignment.}
    \label{fig:module}
\end{figure}

As shown in Figure~\ref{fig:module}, \textsl{AugModule} enhances SAM’s segmentation with an efficient design. It refines LoRA for better storage efficiency without accuracy loss and incorporates prompt information to further improve performance. We first introduce \textsl{SLoRA}, then \textsl{Prompt Augmentation}.

\subsubsection{SLoRA}\label{sec:slora}
LoRA achieves strong performance with few tunable parameters. AsymmLoRA~\cite{asymmetry} further reduces them by freezing matrix~$A$ and fine-tuning only matrix~$B$. This is based on the observation that $A$ is more generalizable, extracting original features into low-rank representations, while $B$ leverages these features to produce the desired output~\cite{asymmetry}.

\begin{figure}
  \centering
  \begin{subfigure}{0.49\linewidth}
    \includegraphics[width=1\linewidth]{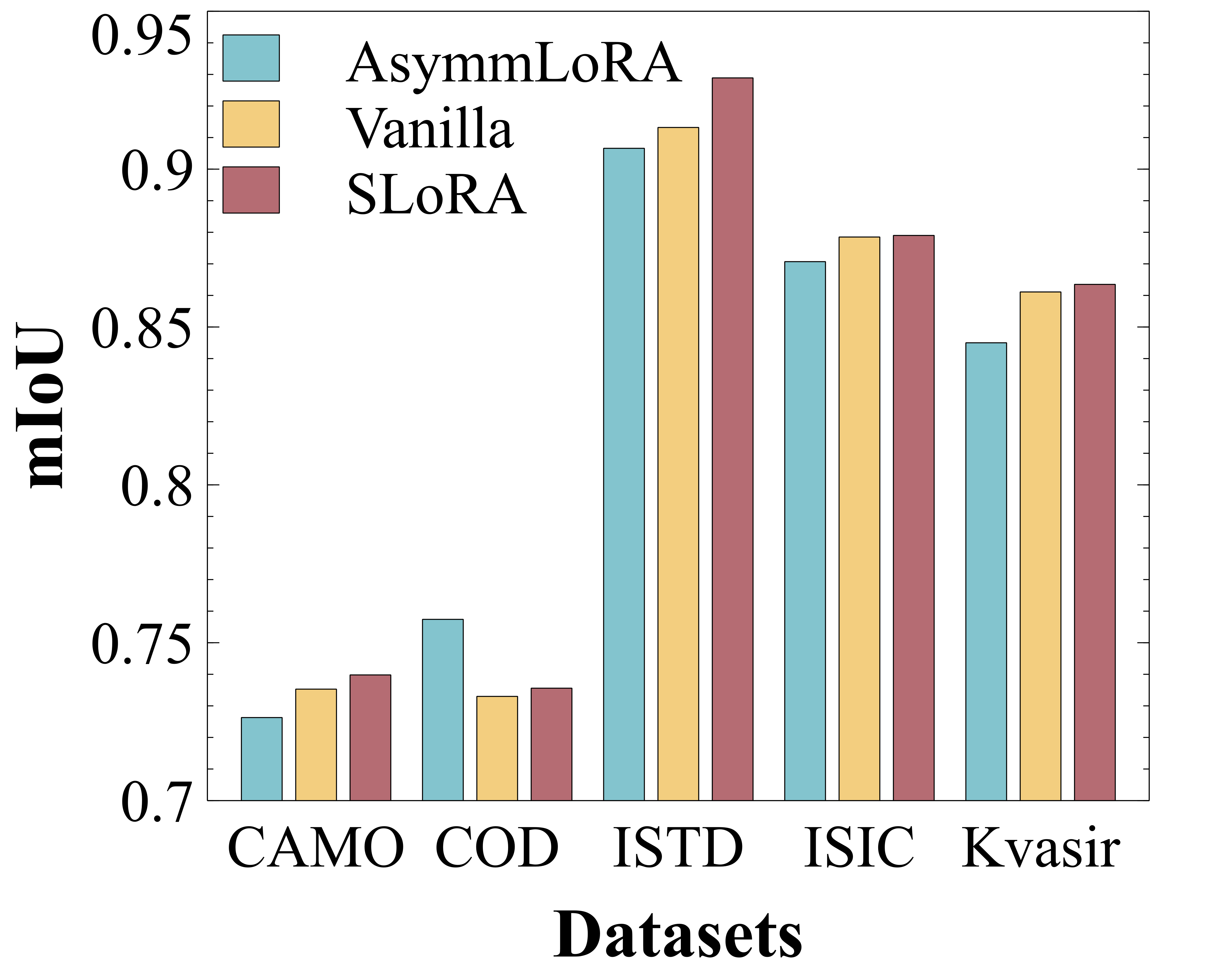}
    \caption{}
    \label{fig:SLora1}
  \end{subfigure}
  \begin{subfigure}{0.49\linewidth}
    \includegraphics[width=1\linewidth]{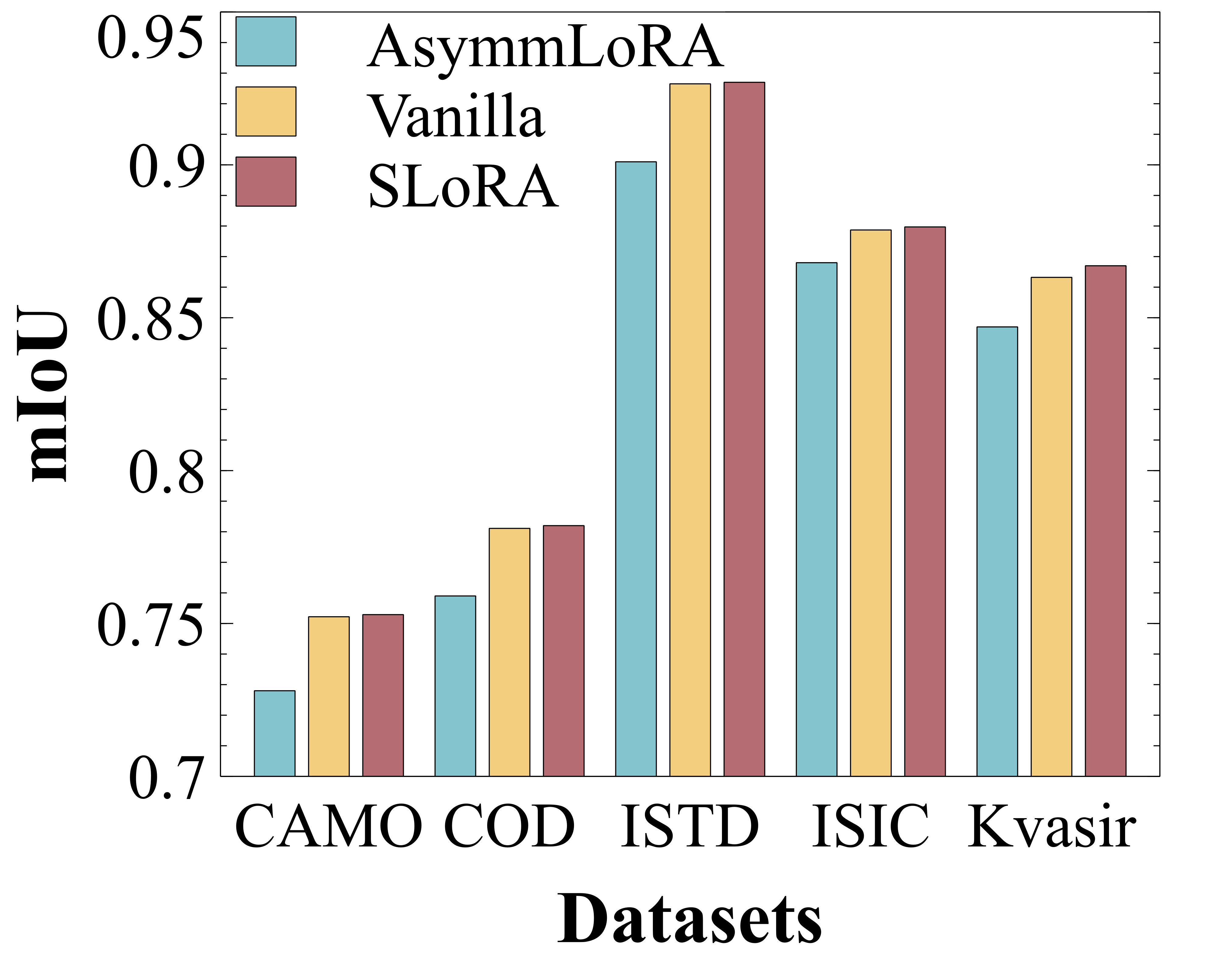}
    \caption{}
    \label{fig:SLora2}
  \end{subfigure}
  \caption{Comparison among AsymmLoRA, vanilla LoRA, and \textsl{SLoRA}. Each method fine-tunes SAM’s image encoder on the CAMO, COD, ISTD, ISIC, and Kvasir datasets for 20 epochs. (a) shows results without \textsl{Prompt Augmentation}, and (b) shows results with \textsl{Prompt Augmentation}.}
  \label{fig:SLora}
\end{figure}

However, this design presents two challenges. \textsl{First}, AsymmLoRA still underperforms vanilla LoRA. As shown in Figure~\ref{fig:SLora1}, we fine-tune the full QKV projection using a single LoRA-type module per attention block. The results show that vanilla LoRA outperforms AsymmLoRA on most datasets and highlight the (albeit small) benefit of training matrix~$A$.
\textsl{Second}, as the number of tasks increases, per-task storage becomes a major bottleneck. Although matrix~$A$ is frozen during training, it still adds to the storage. Therefore, we aim to retain $A$’s functionality while reducing its storage. To address these challenges, we inevitably raise an idea: \textbf{Given matrix~$A$’s generalization capacity, can all matrices~$B$s share a single $A$ while fully leveraging its role?} This insight gives rise to \textsl{SLoRA}, defined as follows:
\begin{equation}
    Y_i = (W_i + B_iA)X_i,
  \label{eq:SLora}
\end{equation}
where $A$ is shared across all $B_i$, and $i$ denotes the index of fine-tuned blocks. As shown in Figure~\ref{fig:SLora1}, \textsl{SLoRA} outperforms both frozen and vanilla LoRA on most datasets. Since \textsl{SLoRA} uses only a single matrix $A$, its total parameter count is nearly half that of vanilla LoRA. Thus, the empirical results demonstrate that \textsl{SLoRA} is an effective and efficient learning module for SAM. Notably, AsymmLoRA performs relatively well on the COD dataset but struggles to integrate prompt information, which we discuss further later.

\subsubsection{Prompt Augmentation}\label{sec:prompt}
Given SAM’s architecture, prompts provide strong guidance for segmentation. Many fine-tuning methods~\cite{SAM-Adapter, SAM-LST} adapt to new tasks by training the mask decoder, but this introduces substantial tunable parameters for each task during CL.
To avoid this, we shift adaptation to the image encoder, bypassing mask decoder training. We use point-type prompts, converting them into heatmaps that integrate with image features. 
Each heatmap is generated by mapping points to its image-space location, assigning intensity values, and applying smoothing for continuity.
We use linear layers for dimensional alignment and inject the resulting \textsl{Prompt Augmentation} into all \textsl{SLoRA}. This brings us to the whole \textsl{AugModule}, as describes below:
\begin{equation}
    Y_i = W_iX_i + B_i(AX_i + P),
  \label{eq:AugAdapter}
\end{equation}
where P represents \textsl{Prompt Augmentation}. We also inject \textsl{Prompt Augmentation} into the MLP (also known as the neck module) in SAM’s image encoder to further enhance its effectiveness, as shown in Figure~\ref{fig:method}. Details of the generating process are provided in the Appendix~B.3.

Comparing Figure~\ref{fig:SLora1} and Figure~\ref{fig:SLora2}, most methods improve with \textsl{Prompt Augmentation}, demonstrating its effectiveness. Both vanilla LoRA and \textsl{SLoRA} outperform AsymmLoRA in Figure~\ref{fig:SLora2}, highlighting the importance of matrix $A$ in extracting low-rank features and integrating information.
The results also show that \textsl{Prompt Augmentation} does not degrade low-rank representations, confirming compatibility LoRA-type methods.
Note that \textsl{Prompt Augmentation} operates in parallel with SAM’s native prompts: box, mask, or point prompts go to the mask decoder, while point prompts (which can be sampled from box or mask prompts) are used for \textsl{Prompt Augmentation}. For simplicity, we use only point prompts throughout this paper.

\subsection{Module Selector}\label{sec:task}
Decoupling incremental knowledge into distinct modules shows strong potential in mitigating forgetting~\cite{PNN}. The main challenge is selecting the appropriate module during testing without access to task identities. This requires a ``selector” that can automatically identify the input domain~\cite{CLSAM, Expert_gate}. However, training such a selector on large buffers of raw images incurs substantial storage costs~\cite{CLSAM}, limiting scalability. To achieve low storage overhead and high performance, the key question is: \textbf{can we leverage SAM’s pre-trained knowledge to aid module selection during testing?} Since SAM’s image encoder extracts rich representations essential for mask decoding, we hypothesize that while it may not perform pixel-wise classification, it can effectively distinguish input domains through class-wise classification.

\begin{figure}[h]
    \centering
    \includegraphics[width=0.9\linewidth]{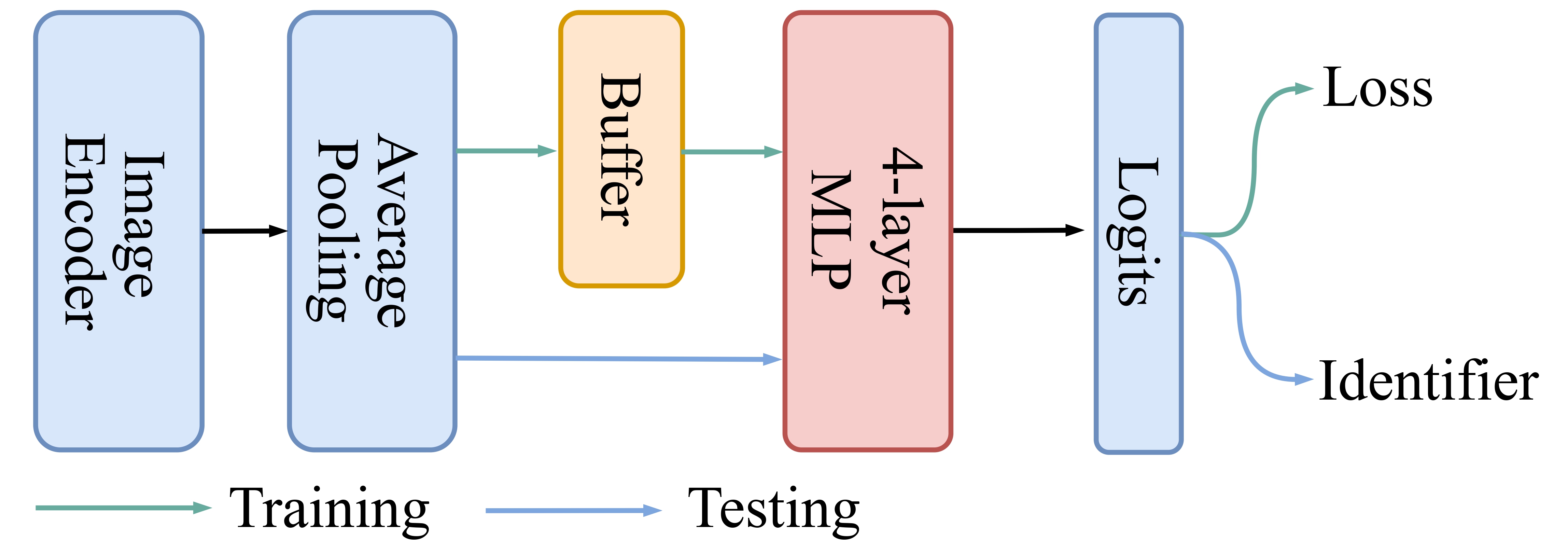}
    \caption{Overview of the \textsl{Module Selector}. We extract a fixed number ($N_e$) of embeddings from a specific block in the image encoder and reduce their dimensions from $(N_e, H, W, D)$ to $(N_e, D)$ by averaging over $H$ and $W$ dimensions. These embeddings ($e_i \in \mathbb{R}^{D}$, where $i = 1, \dots, N_e$) are stored in a buffer aggregating embeddings from all learned domains. The \textsl{Module Selector}, a lightweight MLP with four linear layers, is trained using cross-entropy loss for 25 epochs on this embedding buffer. The layout of the four layers is detailed in the Appendix~B.2.}
    \label{fig:Task1}
\end{figure}
\begin{figure}[h]
    \centering
    \includegraphics[width=1.\linewidth]{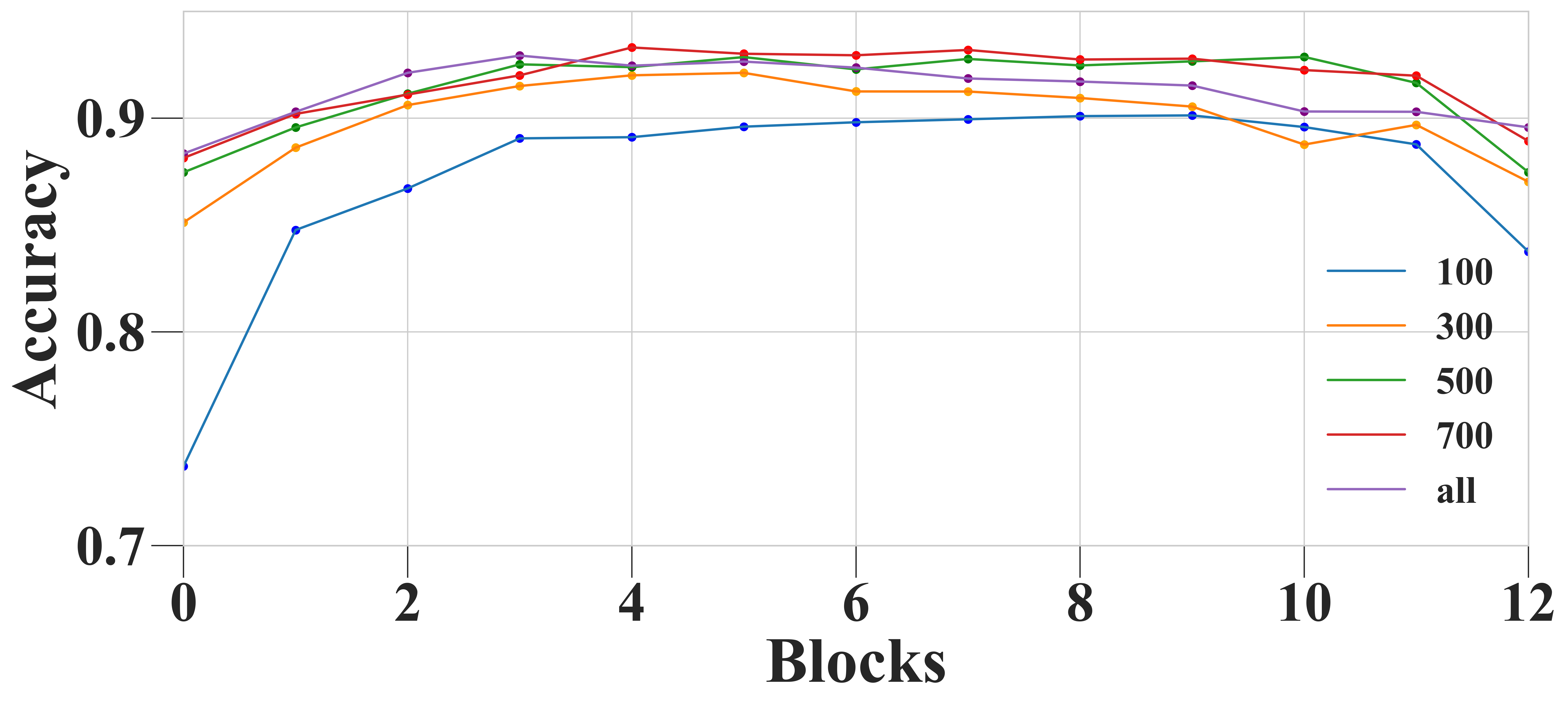}
    \caption{Selection accuracy across blocks in the image encoder of SAM (ViT-b). Each line denotes a different number of stored embeddings per dataset, all showing a similar trend. Results indicate high domain classification accuracy, with middle blocks performing best.}
\label{fig:selecting}
\end{figure}

Inspired by ViT~\cite{Vit}, which uses final embeddings for classification, we leverage latent embeddings from SAM’s image encoder to test our hypothesis. The experimental setup is shown in Figure~\ref{fig:Task1}. Specifically, we divide the entire SAM model $\theta$ into two consecutive parts, $\theta = [\theta_1, \theta_2]$, beginning at a designated intermediate block within image encoder. The embedding extraction is formulated as: 
\begin{equation}
e = f_{\theta_1}(x, \text{Prompt}),
\label{eq:seletor_1}
\end{equation}
where $f$ is the forward function and $x$ is the input image.
As shown in Figure~\ref{fig:selecting}, the model achieves high domain classification accuracy, validating our hypothesis. This motivates the design of the \textsl{Module Selector}, trained on low-consumed embeddings from the buffer rather than raw images, while maintaining high selection accuracy.

At inference time, the appropriate module is selected by:
\begin{equation}
\text{id} = S(e),
\label{eq:selector_2}
\end{equation}
where $S$ denotes the \textsl{Module Selector} and $\text{id}$ is the index of the selected module.
The final SAM inference with the selected module is given by:
\begin{equation}
y = f_{\theta_2^*}(x, \text{Prompt}),
\end{equation}
where $\theta_2^*$ represents $\theta_2$ combined with the selected learning module, and $y$ is the output.

To sum up, let us take an example of the whole process of \method. During training, we obtain embeddings with 300 number per domain from the $6^{th}$ block in the image encoder to train \textsl{Module Selector} and fine-tune only blocks after the $6^{th}$ block by \textsl{AugModule}. 
During inference, we first extract the embedding from the $6^{th}$ block and identify the appropriate module by \textsl{Module Selector}. Then we extract the corresponding \textsl{AugModule} and inject it in the subsequent blocks for inference. 
Note that \method may appear unable to fully recover the original SAM. However, this can be addressed by training the \textsl{Module Selector} on a small COCO subset, effectively representing a virtual module corresponding to the original SAM. In other words, the COCO subset serves as a task in the CL process. The above process and setup are used for SAM, with the SAM2 configuration detailed in the Appendix~B.4. The pseudo-code for \method in both training and inference process is provided in Algorithm~\ref{Algorithm1}.

\begin{algorithm}[ht]
    \caption{\method (Section~4)}
    \label{Algorithm1}
    \begin{algorithmic}[1]
        \STATE \textbf{Input:} SAM, Epoch, Task Number $T$, Training Dataset $D$, Testing Dataset $Test$, Function of Getting Embedding $F()$, Training Function $Train()$, Inference Function $Infer()$, Select Function $Select()$
        
        \STATE Module Set, Buffer $\gets \emptyset, \emptyset$
        
        \FOR{$t=1$ \TO $T$}
            \STATE \textbf{// Training process}
            
            \STATE Model $\gets$ Initialize SAM with a new AugModule
            \STATE Buffer $\gets F(\text{Model}, \text{sample}(D_t))$
            \STATE $Train(\text{Module\_Selector}, \text{Buffer})$
            
            \FOR{epoch=1 \TO Epoch}
                \STATE $Train(\text{Model}, D_t)$  
            \ENDFOR
            
            \STATE Module Set $\gets$ AugModule in Model
            
            \STATE \textbf{// Inference process}
            
            \FOR{data \textbf{in} $Test_t$}
                \STATE embedding $\gets F(\text{SAM}, \text{data})$
                \STATE ID $\gets \text{Module\_Selector}(\text{embedding})$
                \STATE module$_{test} \gets Select(\text{Module Set}, \text{ID})$
                \STATE Model $\gets$ Initialize SAM with module$_{test}$
                \STATE outputs $\gets Infer(\text{Model}, \text{embedding})$
            \ENDFOR
        \ENDFOR
    \end{algorithmic}
\end{algorithm}

\section{Experiment}\label{sec:experiment}
In this section, we evaluate the performance of \method. We begin by outlining the experimental setup, followed by a comparative analysis against existing CL methods and an assessment of robustness. We then conduct ablation studies to validate the effectiveness of \textsl{AugModule} and \textsl{Module Selector}. Additional implementation details and experiments are provided in the Appendix~C and Appendix~D, respectively.

\begin{table*}[h]
\centering
{
\adjustbox{width=\textwidth}{
\begin{tabular}{c|c|ccc|ccc|ccc}
\hline
\multicolumn{11}{c}{Kvasir $\rightarrow$ CAMO $\rightarrow$ ISTD $\rightarrow$ ISIC $\rightarrow$ COD} \\ \hline
\multirow{2}{*}{Method} & \multirow{1}{*}{Storage} & \multicolumn{3}{c|}{\textbf{AA}} & \multicolumn{3}{c|}{\textbf{FM}} & \multicolumn{3}{c}{\textbf{FT}} \\ 
\cline{3-11} & (MB) & mIoU $\uparrow$ & mF1 $\uparrow$ & mMAE $\downarrow$ & mIoU $\downarrow$ & mF1 $\downarrow$ & mMAE $\uparrow$ & mIoU $\uparrow$ & mF1 $\uparrow$ & mMAE $\downarrow$ \\ \hline 
\multicolumn{11}{c}{SAM} \\ 
\hline
\textbf{SAM-Adapter} & 15.72 &0.428& 0.565& 0.125& 0.191& 0.190& -0.046& 0.318& 0.457& 0.196\\

\textbf{SAM-LST} & 44.87 &0.421&	0.533&	0.170&	0.239&	0.222&	-0.180&	0.291&	0.421&	0.221\\

\textbf{AutoSAM} & 158.53&0.184& 0.273& 0.258& 0.330& 0.330& -0.103&	0.124& 0.199& 0.423 \\

\hline

\textbf{LoRA} & 0.30 &0.703&	0.804&	0.064&	0.123&	0.089&	-0.034&	0.513&	0.521&	0.158 \\

\textbf{EWC} & 0.61 &0.716&	0.816&	0.058&	0.111&	0.078&	-0.028&	0.549&	0.663&	0.160 \\
\textbf{ER} & 1125.30 &0.808&	0.881&	0.035&	0.010&	0.007&	-0.003&	0.630&	0.748&	0.087\\
\textbf{DER} & 1125.30 &0.804&	0.879&	0.035&	0.022&	0.015&	-0.005&	0.643&	0.760&	\textbf{0.082} \\
\textbf{SPPA} & 5.24 &0.282&	0.407&	0.149&	0.337&	0.315&	-0.072&	0.417&	0.550&	0.197\\
\textbf{LAG} & 0.61 &0.703& 0.810&	0.063&	0.099&	0.066&	-0.025&	0.452&	0.576&	0.205 \\
\textbf{O-LoRA} & 1.54 &0.704& 0.806& 0.059& 0.091& 0.066& -0.023& 0.519& 0.642& 0.160\\
\textbf{MoDA} & 135.50 &0.756&	0.835&	0.052&	0.020&	0.014&	-0.002&	0.637&	0.757&	0.094\\
\hline
\textbf{SAMCL (Ours)} & 9.11 & \textbf{0.836}& \textbf{0.900}& \textbf{0.029}& \textbf{0.0019}& \textbf{0.0016}& \textbf{-0.0003}& \textbf{0.668}& \textbf{0.773}& 0.089\\ \hline
\multicolumn{11}{c}{SAM2} \\ 
\hline
\textbf{LoRA} & 0.12 &0.664& 0.768& 0.074& 0.169&	0.132& -0.046& 0.479& 0.600& 0.225 \\
\textbf{EWC} & 0.24 &0.697& 0.797&	0.066&	0.139& 0.104& -0.038& 0.502& 0.618&	0.241 \\
\textbf{ER} & 1125.12 &0.801&	0.876& 0.036& 0.028&0.020& -0.008& 0.582& 0.699& \textbf{0.148}\\
\textbf{LAG} & 0.24 &0.646& 0.753&	0.074& 0.182& 0.142& -0.044& 0.477& 0.605& 0.196 \\
\textbf{O-LoRA} & 0.61 &0.655& 0.767& 0.073& 0.146&	0.108& -0.039& 0.499& 0.622& 0.205 \\
\hline
\textbf{SAMCL (Ours)} & 6.12 & \textbf{0.843}& \textbf{0.905}& \textbf{0.028}& \textbf{0.0019}& \textbf{0.0017}& \textbf{-0.0009}& \textbf{0.613}& \textbf{0.722}& 0.149\\ \hline
\end{tabular}}
}
\caption{Comparison results during the CL setting. Considering the compatibility and performance, we select a few representative methods to compare \method with SAM2. The results show the excellent performance of \method. The final testing results for all datasets are provided in the Appendix~D.3.}
\label{tab:main}
\end{table*}

\begin{table*}[h]

\centering
\adjustbox{width=\textwidth}{
\begin{tabular}{c|c||c@{\hspace{2mm}}c@{\hspace{2mm}}c|c@{\hspace{2mm}}c@{\hspace{2mm}}c|c@{\hspace{2mm}}c@{\hspace{2mm}}c|c@{\hspace{2mm}}c@{\hspace{2mm}}c|c@{\hspace{2mm}}c@{\hspace{2mm}}c}
\hline
\multirow{2}{*}{Method} & \multirow{2}{*}{TP (MB)} & \multicolumn{3}{c|}{Kvasir}& \multicolumn{3}{c|}{CAMO} & \multicolumn{3}{c|}{ISTD} & \multicolumn{3}{c|}{ISIC} & \multicolumn{3}{c}{COD} \\ 

\cline{3-17} &&mIoU $\uparrow$ & mF1$\uparrow$ & MAE$\downarrow$ & mIoU$\uparrow$ & mF1$\uparrow$ & MAE $\downarrow$ & mIoU$\uparrow$ & mF1$\uparrow$ & MAE $\downarrow$ &mIoU $\uparrow$ & mF1$\uparrow$ & MAE$\downarrow$ &mIoU $\uparrow$ & mF1$\uparrow$ & MAE$\downarrow$ \\ \hline
\multicolumn{17}{c}{SAM} \\ \hline

\textbf{SAM} & 0& 0.736& 0.824& 0.082& 0.579& 0.701& 0.111& 0.612& 0.723& 0.091& 0.650&	0.761& 0.160& 0.655& 0.763&	0.042\\ \hline

\textbf{SAM-Adapter} &15.724& 0.561& 0.720& 0.098& 0.532& 0.709& 0.114& 0.824& 0.897& 0.038& 0.718& 0.841& 0.050& 0.513& 0.673& 0.062\\
\textbf{SAM-LST} & 44.874& 0.709& 0.817& 0.050& 0.471& 0.624& 0.137& 0.854&	0.904&	0.028& 0.826&	0.898&	0.051& 0.510&	0.652&	0.057\\

\textbf{AutoSAM} & 158.532 & 0.634& 0.726&	0.080& 0.261& 0.379& 0.334& 0.697& 0.756&	0.087& 0.727& 0.805& 0.097& 0.255& 0.355& 0.173 \\ 

\textbf{LoRA} & 0.308& 0.861&	0.919&	\textbf{0.022}& 0.735&	0.832&	0.055& 0.913& 0.947&	0.013& 0.878& 0.933& 0.035& 0.733&	0.829&	0.026\\ 
\textbf{AsymmLoRA} & 0.308& 0.847& 0.909& 0.028& 0.728& 0.829& 0.059& 0.901& 0.940& 0.016& 0.868& 0.927& 0.037& 0.759& 0.847& 0.023\\
\textbf{HydraLoRA} & 0.674& 0.863&	0.922&	\textbf{0.022}&	0.743&	0.837&	0.052&	0.923&	0.955&	0.010&	0.875&	0.931&	0.036&	0.698&	0.802&	0.030 \\
\hline
\textbf{SLoRA} & 0.190&	0.859&	0.916& 0.025& 0.738& 0.836&	0.052&	0.916&	0.950&	0.012&	0.879&	0.931&	0.036&	0.735&	0.832&	0.025\\
\textbf{AugModule} & 0.233 & \textbf{0.867}& \textbf{0.922}& 0.023& \textbf{0.751}& \textbf{0.845}& \textbf{0.051}& \textbf{0.927}& \textbf{0.957}& \textbf{0.009}& \textbf{0.879}& \textbf{0.933}& \textbf{0.034}& \textbf{0.782}& \textbf{0.864}& \textbf{0.019}\\ 
\hline
\multicolumn{17}{c}{SAM2} \\ 
\hline
\textbf{SAM2} & 0& 0.767& 0.844& 0.081&	0.656&	0.760&	0.111&	0.522&	0.637&	0.200&	0.615&	0.724&	0.232&	0.690&	0.788&	0.042\\ 
\hline
\textbf{LoRA} & 0.123& 0.860&	0.918&	0.023&	0.743&	0.837&	0.052&	\textbf{0.934}&	\textbf{0.962}&	\textbf{0.009}& 0.874& 0.931& 0.034&	0.752&	0.843&	0.023\\
\textbf{AsymmLoRA} & 0.123& 0.864& 0.920& 0.023& 0.750& 0.845& 0.055& 0.931& 0.959& 0.011& 0.877& 0.932& 0.034& 0.750& 0.841& 0.024\\ 
\textbf{HydraLoRA} & 0.273&	0.860&	0.916&	0.029&	0.757&	0.848&	0.049&	0.931&	0.958&	0.011& 0.880&	0.933&	0.034&	0.781&	0.863&	0.020\\ 
\hline
\textbf{SLoRA} & 0.079& 0.872&	0.927&	0.022&	0.752&	0.843&	0.050&	0.930& 0.953&	0.0013&	0.873&	0.929&	0.0335&	0.751&	0.843&	0.024\\
\textbf{AugModule} & 0.079 & \textbf{0.884}& \textbf{0.935}& \textbf{0.015}&	\textbf{0.779}&	\textbf{0.866}&	\textbf{0.044}&	\textbf{0.934}&	0.960&	\textbf{0.009}&	\textbf{0.881}&	\textbf{0.935}&	\textbf{0.033}&	\textbf{0.786}&	\textbf{0.868}&	\textbf{0.019}\\  
\hline
\end{tabular}}
\caption{Comparison with fine-tuning methods over $20$ epochs. We selected three fine-tuning methods for SAM: SAM-Adapter, SAM-LST, and AutoSAM. For LoRA-type methods (LoRA, AsymmLoRA, and HydraLoRA), we set the rank to $10$ for both SAM and SAM2, consistent with \textsl{AugModule} and \textsl{SLoRA} (without \textsl{Prompt Augmentation}). Additionally, for HydraLoRA, we configured the number of multiple matrices $B$s to $4$.}
\label{tab:abla_augmodule}
\end{table*}

\subsection{Experimental setting}\label{sec:setting}
For datasets, given limitations noted in prior work~\cite{notperfect}, we choose three challenging domains: camouflaged, shadow, and medical domains. For camouflaged, we use COD~\cite{COD} and CAMO~\cite{CAMO}; for medical, ISIC~\cite{ISIC} and Kvasir~\cite{Kvasir}; and for shadow, ISTD~\cite{ISTD}. The CL dataset order used in the main paper is Kvasir, CAMO, ISTD, ISIC, and COD, interleaving different domains. All datasets are trained for 20 epochs and evaluated by mIoU, mF1, and mMAE. We adopt AA, FM, and FT for CL evaluation. We measure storage efficiency by accounting for all extra storage consumption incurred during CL, excluding the base SAM model. To support reverting \method to the original SAM, we train \textsl{Module Selector} on selected $100$ samples from COCO~\cite{COCO} after CL.

We compare the following representative methods:
(1) SAM fine-tuning (SAM-Adapter~\cite{SAM-Adapter}, SAM-LST~\cite{SAM-LST}, AutoSAM~\cite{AutoSAM});
(2) General CL methods (EWC~\cite{EWC}, ER~\cite{ER}, DER~\cite{DER});
(3) Architecture-based method for foundation models (O-LoRA~\cite{O-Lora});
(4) Continual segmentation (SPPA~\cite{SPPA}, LAG~\cite{LAG});
(5) SAM-specific continual learning (MoDA~\cite{CLSAM} + HQ-SAM~\cite{sam_hq}).
For fair comparison, all general CL and segmentation methods are integrated with LoRA. We use the ViT-b version of SAM and the tiny version of SAM2 (v2.1), selecting three random object points as prompts during training and inference. The batch size for SAM is $8$, whereas for SAM2 it is $16$.

For \method, we use AdamW optimizer with an initial learning rate of 0.005 and cosine decay. Each task stores 300 embeddings, same as ER and DER. For MoDA, we follow its original setup, storing 10 raw samples per task. The low-rank dimension of \textsl{SLoRA} is $10$, consistent across all LoRA-based baselines. \textsl{Module Selector} is trained with cross-entropy loss $\mathscr{L}_c$.
For fine-tuning SAM with \textsl{AugModule}, we follow the official SAM implementation~\cite{sam}, modifying only loss coefficients. The IoU head is trained with MSE loss to measure the gap between the predicted and true IoU. The mask prediction is supervised by a weighted sum of focal~\cite{focalloss} and dice loss~\cite{diceloss}. The full loss function is:
\begin{equation}
    \mathscr{L}_s = \mathcal{L}_{Focal} + 10 * \mathcal{L}_{Dice} + \mathcal{L}_{Mse}.
  \label{eq:Loss_s}
\end{equation}

\subsection{Comprehensive Results}\label{sec:main}
As shown in Table~\ref{tab:main}, \method with SAM and SAM2 demonstrates exceptional performance during CL. The AA and FM metrics demonstrate that \method achieves strong average accuracy while effectively mitigating forgetting. Notably, the FT metric reveals that \method exhibits excellent transferability without training on unseen tasks. 

In contrast, fine-tuning methods (SAM-Adapter, SAM-LST, and AutoSAM) perform poorly in the CL setting, underscoring the need to enhance SAM with CL capabilities across different domains. The FT value also suggests overfitting in fine-tuning methods, indicating a lack of generalization to unseen domains. Additionally, continual segmentation methods (SPPA and LAG) struggle to delay forgetting when integrated with SAM. 

General CL methods (EWC, ER, and DER) can delay forgetting to some extent, particularly the replay-based methods ER and DER, as indicated by the FM metric. However, \method outperforms these approaches by effectively preventing interference between different knowledge domains. Moreover, ER and DER exhibit low efficiency, storing 300 raw image samples ($x^i \in \mathbb{R}^{3\times 256\times 256}$) per domain in buffer, totaling approximately $225$MB. In contrast, \method stores an equal number of embedding samples per domain ($e^i \in \mathbb{R}^{768}$ for SAM and $e^i \in \mathbb{R}^{384}$ for SAM2), resulting in a storage cost of only 0.878 MB for SAM and 0.439 MB for SAM2—representing a remarkable reduction of up to $256\times$ and $512\times$, respectively. The total storage cost is reduced by 123.52$\times$ and 183.84$\times$ for SAM and SAM2 compared with ER and DER, respectively, as shown in Table~\ref{tab:main}.
Notably, while ER and DER require training on both previous and new samples during CL, \method only trains on stored embeddings once using a simple MLP within $25$ epochs before addressing a new task, resulting in a negligible training burden. 
Although some methods, such as EWC and O-LoRA, incur lower storage costs, their performance is significantly inferior to that of \method and ER.

Compared to the architecture-based method O-LoRA, \method excels, as O-LoRA fails to achieve true orthogonality during testing. Additionally, although MoDA is an existing CL method for SAM, it performs worse than \method. Following its original design, MoDA requires storing $10$ samples per domain in a buffer, resulting in a storage cost that is still $8.53\times$ higher than that of \method. The total storage consumption reaches over $14.87\times$ that of \method in SAM, as shown in Table~\ref{tab:main}.

Therefore, this comparison underscores the superior storage efficiency and CL performance of \method with SAM, highlighting its significance in the field. The visualized result during CL process is provided in Appendix~D.2.



\subsection{Robustness Analysis}\label{sec:robust}
\begin{figure}[ht]
    \centering
    \includegraphics[width=0.9\linewidth]{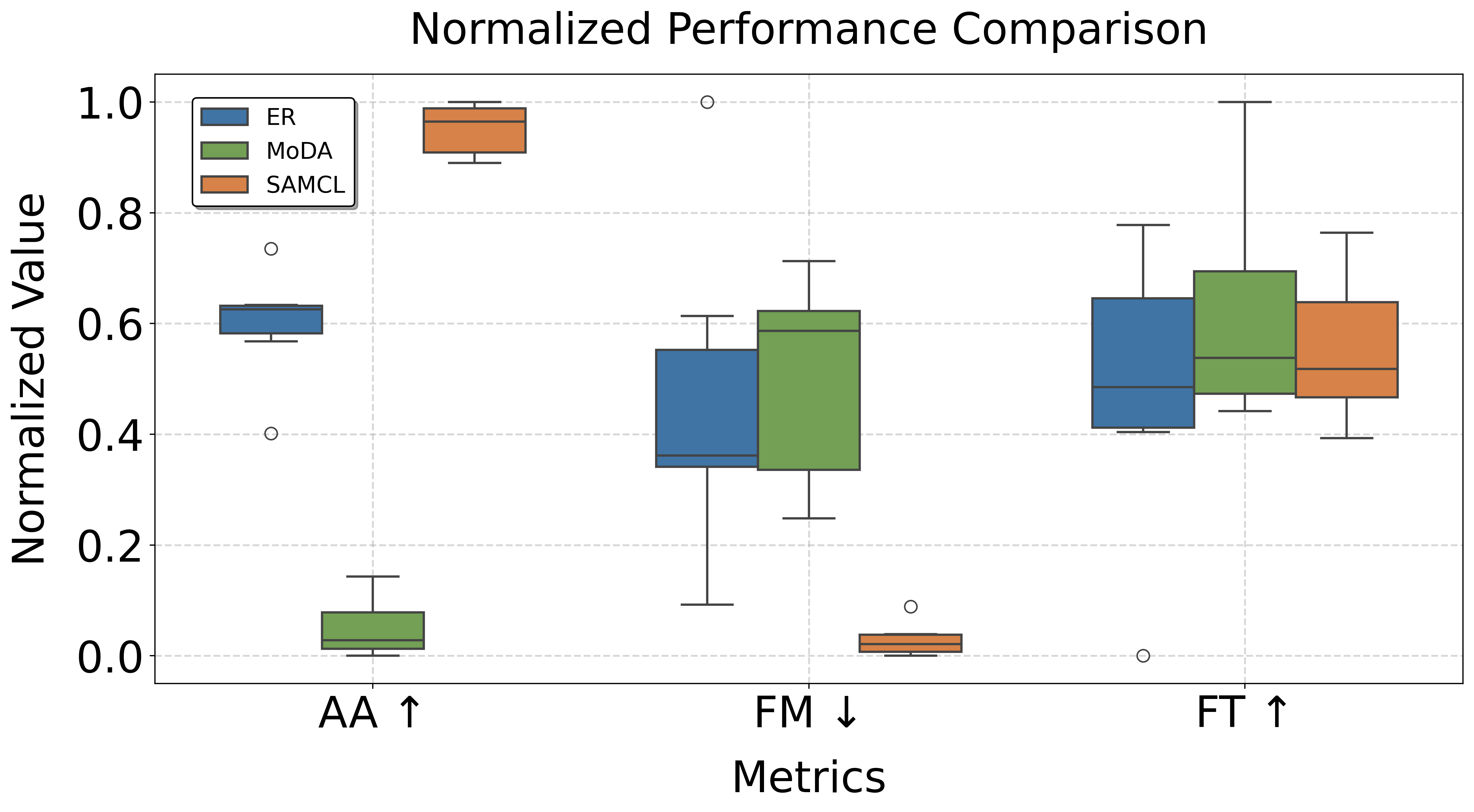}
    \caption{Analysis of the robustness. Following the settings outlined in Table~\ref{tab:main}, we conduct experiments across six different orders. For clarity, we normalize the mIoU metric values.}
    \label{fig:robust}
\end{figure}

CL is challenged not only by catastrophic forgetting but also by the sensitivity to task order~\cite{order}, which can lead to unstable performance. This section evaluates the performance of \method under different task orderings. In Figure~\ref{fig:robust}, we compare the robustness of \method with ER and MoDA. The results on AA and FM show that \method achieves superior robustness in both learning and retention. This is because \method allocates different knowledge to distinct learning modules, and the accurate selector in SAMCL chooses the most appropriate one.
For transferring, all three methods exhibit similar levels of dispersion. Detailed values and comparisons for each order of these two methods are presented in Appendix~D.3.

\subsection{Ablation Study}\label{sec:ablation}
This section displays ablation studies about \textsl{AugModule} and \textsl{Module Selector} in \method. We firstly presents the overall ablation study of these two components, as shown in Table~\ref{tab:ablation}. Based on this experiment, we will provide a detailed explanation of these two components.

\newcommand{\cmark}{\checkmark} 
\newcommand{\xmark}{\text{\sffamily X}}
\begin{table}[h]
\centering 
\scalebox{0.9}{
\begin{tabular}{c@{\hspace{2mm}}c@{\hspace{2mm}}|ccc}
\toprule 
 \textbf{AugModule} &\textbf{Module Selector} & \textbf{AA} $\uparrow$ & \textbf{FM} $\downarrow$ & \textbf{FT} $\uparrow$  \\
\midrule
$\times$ & $\times$ & 0.703  & 0.123 & 0.513 \\
\cmark & $\times$ &  0.748& 0.086& 0.570\\
\cmark & \cmark & \textbf{0.836}& \textbf{0.0019}& \textbf{0.668}\\
\bottomrule
\end{tabular}
}
\caption{Ablation study of all components in \method with SAM. The experiment setting is the same as in Table~\ref{tab:main} and only uses mIoU metrics for evaluation. We use LoRA as a baseline reference without \method, shown in the first line. }
\label{tab:ablation}
\end{table}

\subsubsection{Discussion on AugModule}\label{sec:ablation_augadapter} 
The \textsl{AugModule}, including \textsl{SLoRA} and \textsl{Prompt Augmentation}, is designed to learn from a new domain, characterized by its strong learning ability and low storage cost, both essential for CL. In this section, we evaluate these two aspects. As shown in Table~\ref{tab:abla_augmodule}, we compare \textsl{AugModule} with other fine-tuning methods in both SAM and SAM2. The results show that \textsl{AugModule} achieves great performance in learning all domains over $20$ epochs. The cost of storing parameters is significantly lower compared to other methods. 

For LoRA-type methods, although AsymmLoRA reduces the number of fine-tuning parameters, its storage cost remains the same as LoRA, and its performance is inferior. \textsl{SLoRA}, we proposed, uses almost half of the storage costs with similar performance compared with LoRA and outperforms AsynmmLoRA. HydraLoRA achieves better performance across most datasets than LoRA and \textsl{SLoRA}. However, \textsl{AugModule}, with the help of \textsl{Prompt Augmentation}, outperforms HydraLoRA. Meanwhile, for SAM, \textsl{AugModule} uses only $0.233$ MB, representing a reduction of $24.3\%$, compared with LoRA. For SAM2, it utilizes just $0.079$ MB, resulting in a $35.7\%$ reduction, compared with LoRA. Therefore, these results show the both effectiveness and storage efficiency.
Due to the poor performance of other fine-tuning methods (SAM-Adapter, SAM-LST, and AutoSAM) for SAM, we provide analysis in the Appendix~D.6.

\subsubsection{Exploration on Module Selector}\label{sec:ablation_classifier}
\textsl{Module Selector} in \method is responsible for selecting appropriate modules for inference. This function not only mitigates the forgetting problem but also facilitates transfer to unseen domains. As shown in Table~\ref{tab:ablation}, there is a minimal difference in the FM metric when only \textsl{AugModule} is used. Surprisingly, after applying \textsl{Module Selector}, the FM score significantly decreases while the FT score greatly increases, demonstrating its effectiveness in both mitigating forgetting and enhancing transferability. In this section, we verify these two factors of \textsl{Module Selector}.

\begin{figure}[h]
    \centering
    \includegraphics[width=0.95\linewidth]{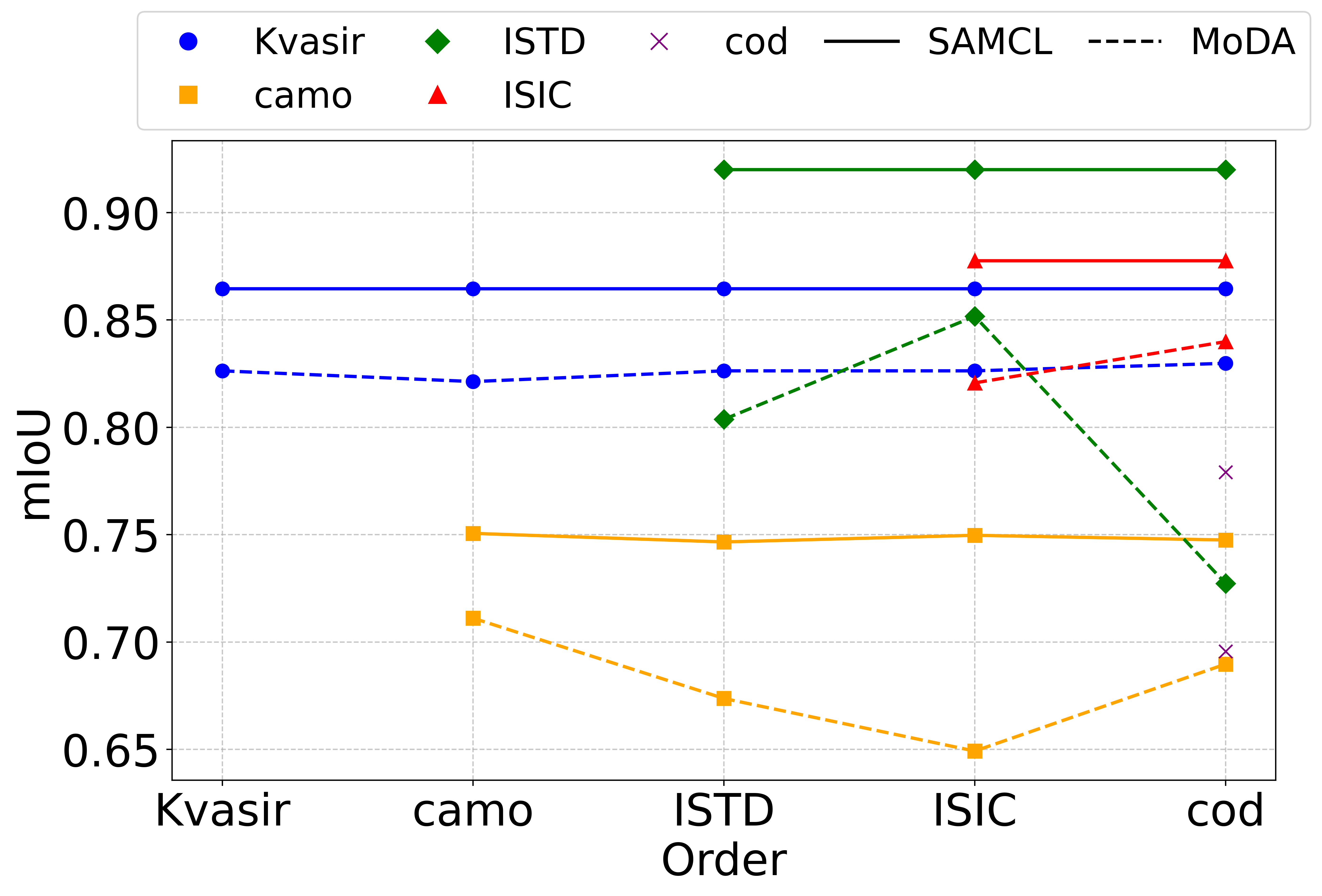}
    \caption{Comparison of performance changes between \method (solid line) and MoDA (dashed line) across each domain during CL. The dataset order and experimental setup are consistent with Table~\ref{tab:main}.}
    \label{fig:combined_selector}
\end{figure}

To evaluate the effectiveness of mitigating forgetting, we conduct experiments that gradually add modules, starting with only the \textsl{Module Selector}. As shown in Figure~\ref{fig:combined_selector}, we compare \method with MoDA, an existing CL method for SAM that employs a similar selector. Compared to MoDA, \method exhibits greater stability in mitigating forgetting, with minimal changes across each domain during CL. This is primarily due to the \textsl{Module Selector}'s exceptional ability to select appropriate modules during the testing process.

\begin{figure}[h]
  \centering
  \begin{subfigure}{0.53\linewidth}
    \includegraphics[width=1\linewidth]{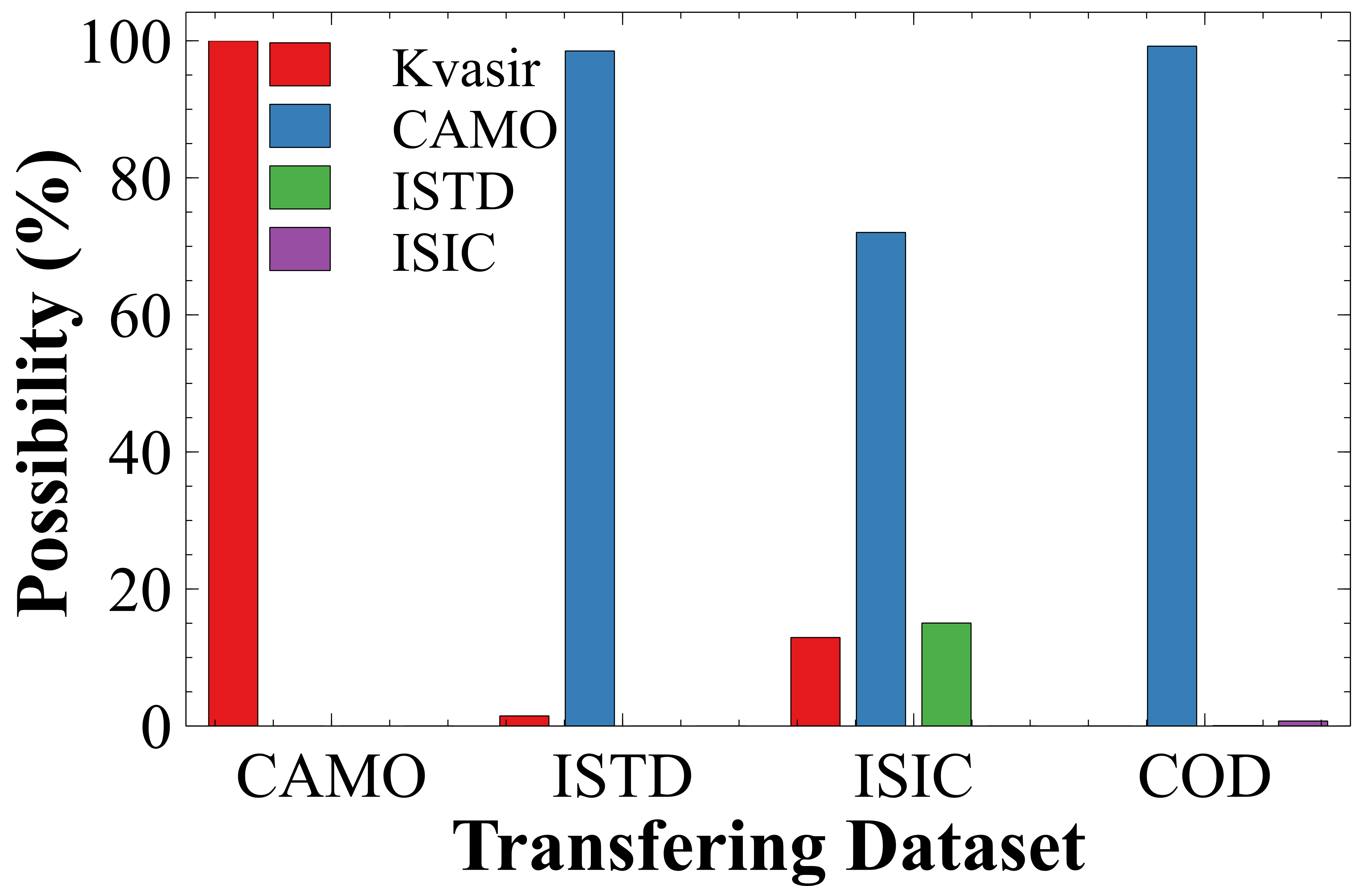}
    \caption{}
    \label{fig:possibility}
  \end{subfigure}
  \begin{subfigure}{0.45\linewidth}
    \includegraphics[width=1\linewidth]{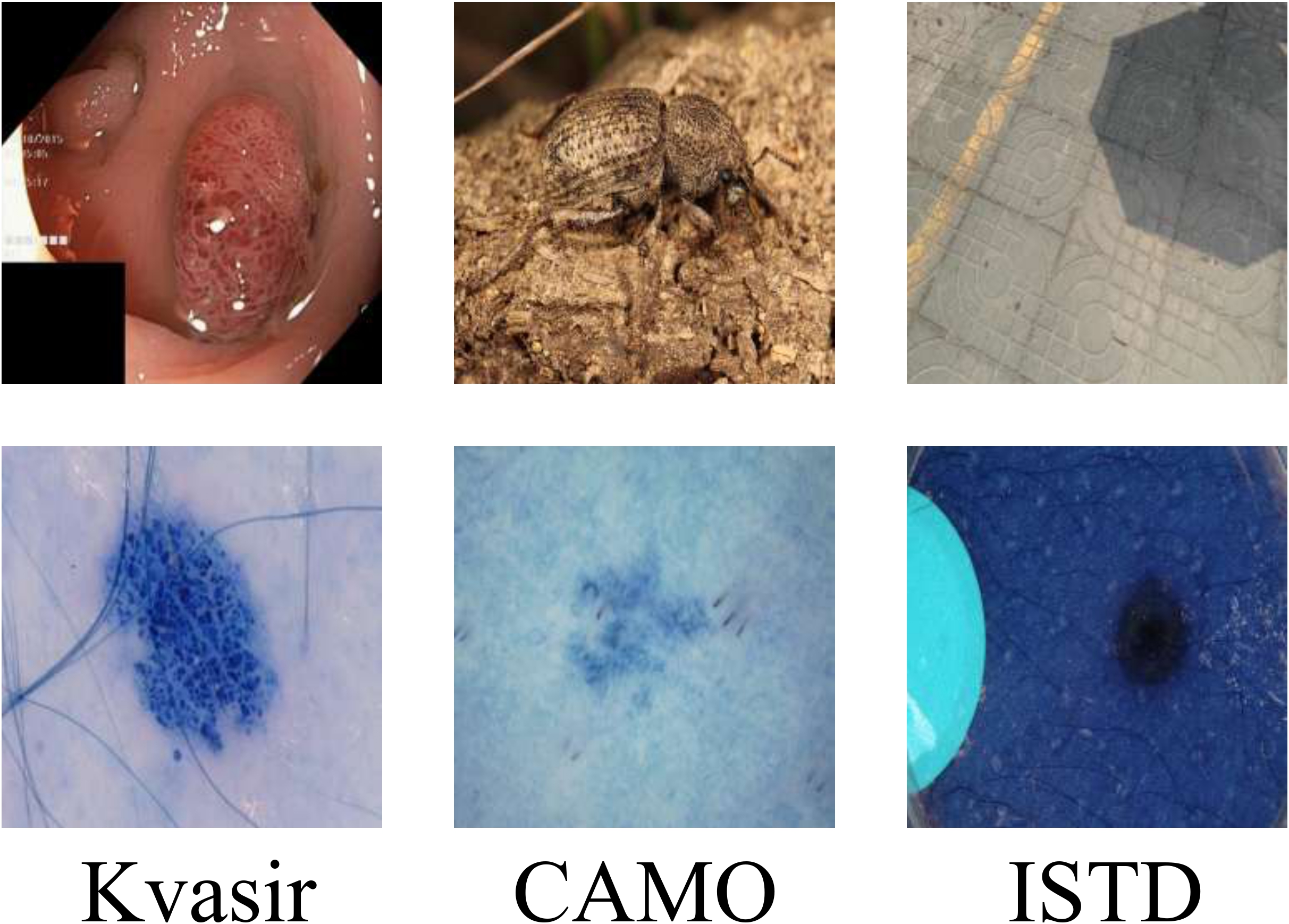}
    \caption{}
    \label{fig:transfer}
  \end{subfigure}
  \caption{Transferring ability of \textsl{Module Selector}. The order of datasets and experimental setup are the same as in Table~\ref{tab:main}. (a) shows the possibility of selecting modules for testing unseen domains. The x-axis represents testing unseen domains, while the y-axis indicates the possibility of utilizing seen domains for unseen domains. (b) provides details of selecting different learned domains (including Kvasir, CAMO, and ISTD) for testing the ISIC dataset. The first row displays examples from the seen domains, and the second row lists examples from the ISIC dataset that are recognized as belonging to specific seen domains.}
  \label{fig:transferring}
\end{figure}

To verify the effectiveness of transferability, we conduct an experiment to show the potential of using previously seen domains to test unseen domains. To better illustrate the transferability across learning domains, we evaluate only the relationships between learned domains and do not assess the reversion to the original SAM. As illustrated in Figure~\ref{fig:possibility}, the results show that the unseen domain can benefit from multiple previous domains. For example, when testing on the unseen COD dataset, the \textsl{Module Selector} automatically selects the module trained on the CAMO dataset. This is primarily because both datasets belong to the camouflage domain.
To further verify its effectiveness, we examine the selection similarity during testing on the ISIC dataset, shown in Figure~\ref{fig:transfer}. For example, the first image in the second row is automatically selected as the Kvasir domain, as its texture resembles that of objects in Kvasir, compared to the first image in the first row. The object in the second image appears in a camouflaged environment, so it is automatically identified as belonging to the CAMO domain. The third image, resembling a shadow object, is identified as part of the ISTD domain. Therefore, with the help of the \textsl{Module Selector}, \method effectively leverages all previously learned knowledge to address unseen domains, showcasing its excellent transferability.

\section{Conclusion}\label{sec:conclusion}
This paper proposes an efficient CL method, termed \method, to enable SAM to learn continually across dynamic domains by \textsl{AugModule} and \textsl{Module Selector}. Crucially, by exploring and reducing the redundancy in module design and selection in SAM, \method achieves a more effective approach to delay \textit{catastrophic forgetting} with extreme storage efficiency. In summary, we believe that \method will stimulate significant interest in SAM applications and provide valuable insights for future research.
\section*{Acknowledgements}
The work was supported by the Fundamental Research Funds for the Central Universities (No. XJSJ25005), Ministry of Education Top-notch Student Training Program in Basic Disciplines 2.0 Research Topics (No. 20252012),  the Natural Science Basis Research Plan in Shaanxi Province of China (No. 2025JC-JCQN-089) and the Outstanding Youth Science Foundation of Shaanxi Province under Grant 2025JC-JCQN-083. Thanks to the help provided by the National Experimental Teaching Demonstration Center for Computer Network and Information Security affiliated with Xidian University.

\bibliography{aaai2026}

\appendix
\newpage

\section{Introduction}\label{sup_sec:introduction}
This supplementary material provides additional details of our method. The remainder of this document is organized as follows:
\begin{itemize}
    \item \textbf{Appendix~\ref{sup_sec:method_details}} presents a comprehensive details of \method for both SAM and SAM2.
    \item \textbf{Appendix~\ref{sup_sec:experiment_implementation_details}} outlines further experimental implementing details.
    \item \textbf{Appendix~\ref{sup_sec:further_experiments}} reports additional results to further demonstrate the effectiveness of our approach.
    \item \textbf{Appendix~\ref{sup_sec:limitations_and_future_works}} discuss limitations and future directions.
\end{itemize} 

\section{Method Details}\label{sup_sec:method_details}
In this section, we first describe the inference process (Appendix~\ref{sup_sec:testing_process}). We then present the layout of the \textsl{Module Selector} (Appendix~\ref{sec:layout_of_module_selector}) and the generating process of \textsl{Prompt Augmentation} (Appendix~\ref{sup_sec:details_of_prompt_augmentation}), both applicable to SAM and SAM2. While the main paper focuses on \method with SAM, we conclude by introducing its extension to SAM2 (Appendix~\ref{sup_sec:samcl_with_sam2}).

\subsection{Inference Process}
\label{sup_sec:testing_process}

\begin{figure*}[htbp]
    \centering
    \includegraphics[width=1\linewidth]{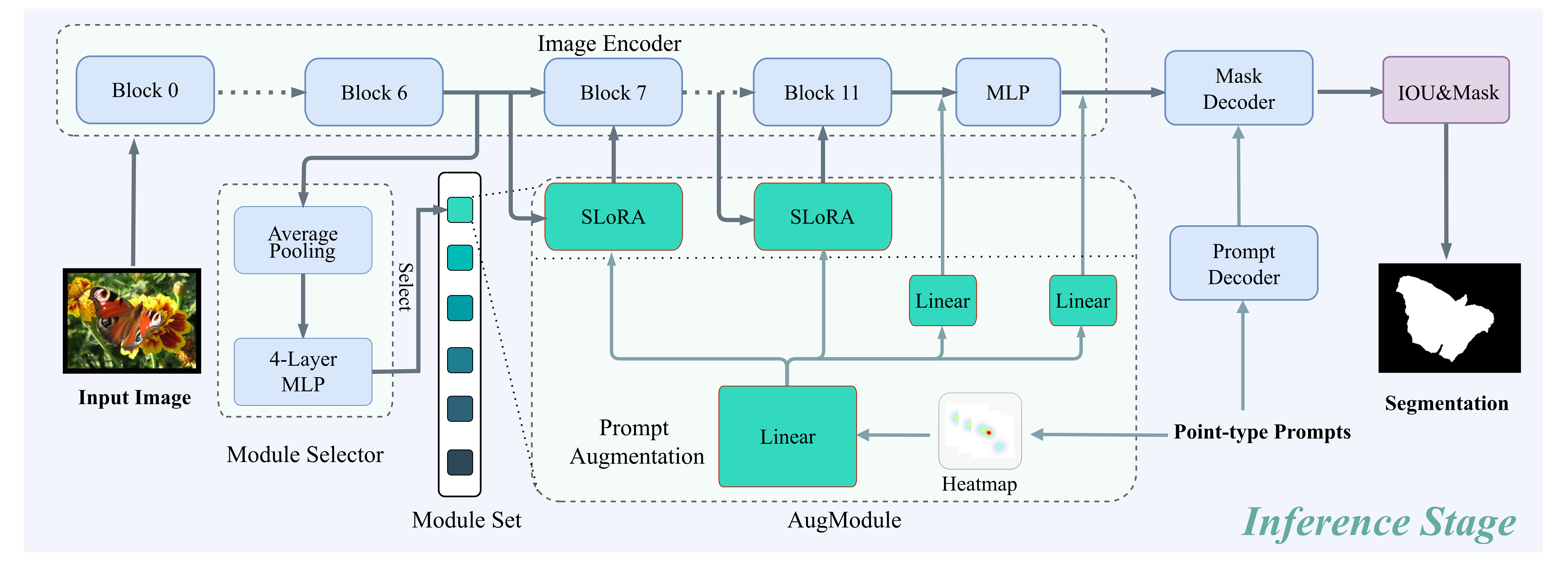}
    \caption{Overview of SAMCL with SAM in the inference process. Firstly, the latent embedding is obtained, and then \textsl{Module Selector} selects the appropriate module to be integrated into the subsequent blocks. Notably, prediction continues after combining SAM with the selected module rather than starting from scratch. }
    \label{fig:method_test}
\end{figure*}

As illustrated in Figure~\ref{fig:method_test}, the inference process of \method is detailed step-by-step. 

\subsection{Layout of Module Selector}\label{sec:layout_of_module_selector}
In this part, we display the layout of \textsl{Module Selector}. \textsl{Module Selector} is an MLP consisting of four linear layers, denoted as $l_1$ to $l_4$. we configure the linear layers as follows: $l_1 \in \mathbb{R}^{D \times D}$, $l_2 \in \mathbb{R}^{D \times D//4}$, $l_3 \in \mathbb{R}^{D//4 \times D//4}$, and $l_4 \in \mathbb{R}^{D//4 \times T}$ ($D$ represents dimension of the latent embedding and $T$ represents the number of incremental domains). Using the experimental setup of SAM described in Section~\ref{sec:experiment} as an example, the embedding dimension is $(N_e, 64, 64, 768)$, where $N_e$ represents the number of stored embeddings per domain. After applying average pooling, the resulting dimension is $(N_e, 768)$. Consequently, we configure the linear layers as follows: $l_1 \in \mathbb{R}^{768 \times 768}$, $l_2 \in \mathbb{R}^{768 \times 192}$, $l_3 \in \mathbb{R}^{192 \times 192}$, and $l_4 \in \mathbb{R}^{192 \times 6}$ ($6$ is represents the six incremental domains). The total resource consumption is approximately $2.95$ MB, which serves as a shared component across all incremental datasets. \textsl{Module Selector} with SAM2 has a similar structure, differing only in dimensions.

\subsection{Generating Process of Prompt Augmentation}\label{sup_sec:details_of_prompt_augmentation}
This part describes the heatmap generation process for \textsl{Prompt Augmentation}. We first apply Gaussian blur with standard deviation $\sigma=9$ along both x and y axes to create smoothed heatmaps of dimensions $(batch, 1, 64, 64)$, where $batch$ denotes batch size. The maps are then expanded to $(batch, r, 64, 64)$ through linear projection along the second dimension, where $r$ corresponds to the low-rank dimension in \textsl{SLoRA}. These processed heatmaps are subsequently integrated into \textsl{SLoRA} as augmentation prompts.

\subsection{SAMCL with SAM2}\label{sup_sec:samcl_with_sam2}
Unlike SAM, SAM2 \cite{sam2} incorporates memory modules for video segmentation, which are unnecessary for image segmentation. As a result, SAM2 shows significant similarity in image segmentation tasks. Both of them use image encoder, prompt encoder, and mask decoder for image segmentation. The primary distinction lies in the image encoder model used: SAM employs the ViT model~\cite{Vit}, while SAM2 utilizes the Hiera model~\cite{Hiera}. 

To incorporate \method into SAM2, two primary questions arise: \textbf{(1) Can \textsl{AugModule} achieve similar performance improvements in learning from the new domain? (2) Can the image encoder in SAM2 effectively distinguish input images from different domains, enabling the \textsl{Module Selector} to accurately select appropriate modules during inference?} Addressing these questions is crucial for the success of \method with SAM2 in learning, maintaining, and transferring knowledge during CL.

To address the first question, fine-tuning accuracy across different domains serves as an effective measure to evaluate the performance of \textsl{AugModule}. As shown in Table~\ref{tab:abla_augmodule}, \method with SAM2 demonstrates exceptional performance in all tested domains while reducing parameter consumption to 35.7\% compared to vanilla LoRA with SAM2. Furthermore, due to the enhancements in SAM2 over SAM, \method with SAM2 outperforms \method with SAM. Same to \method with SAM, \method with SAM2 also fine-tune last blocks in the image encoder. The number of fine-tuning blocks will be discussed later.

\begin{figure}[h]
    \centering
    \includegraphics[width=0.85\linewidth]{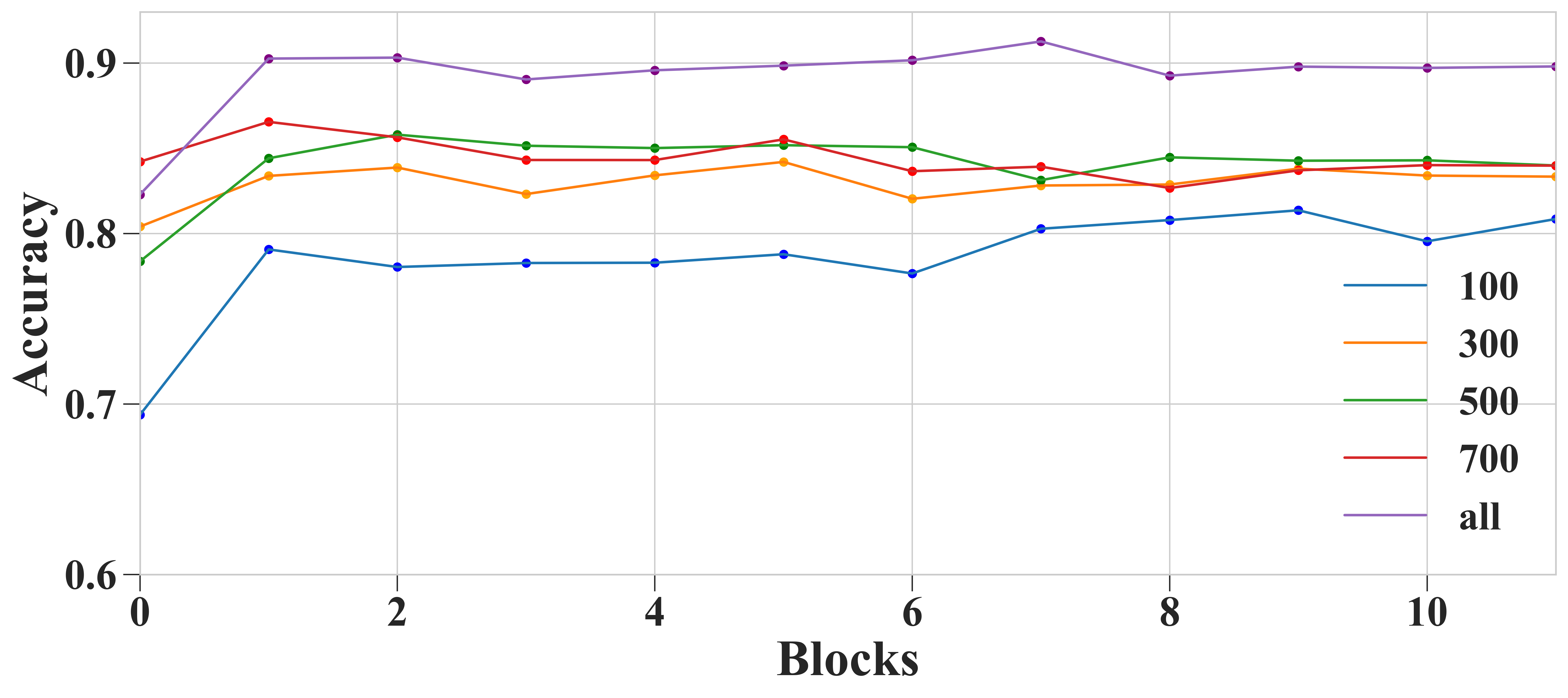}
    \caption{The selecting ability of each block in the image encoder of the tiny version of SAM2. Module Selector is a four-layer MLP, similar to \textsl{Module Selector} for SAM. 
    Different lines represent varying numbers of stored embeddings for each dataset. All lines exhibit a similar trend. The result shows high accuracy in distinguishing between different domains within 25 epochs, with the middle block achieving particularly high performance.}
\label{fig:selecting_sam2}
\end{figure}

To address the second question, we conduct a similar experiment in the main paper to evaluate the selecting ability in the image encoder of SAM2. As shown in Figure~\ref{fig:selecting_sam2}, experiments demonstrate high accuracy in distinguishing domains, which is same to the SAM. Therefore, \textsl{Module Selector} is still suitable for SAM2.

Based on the previous validations, \method is suitable for SAM2. The entire process for implementing \method with the tiny version of SAM2 is as follows. We extract $300$ embeddings per domain from the 5$^{th}$ block of the image encoder to train the \textsl{Module Selector} and use \textsl{AugModule} to fine-tune only the blocks after the 5$^{th}$ block. During inference, we first extract the embedding from the 5$^{th}$ block and identify the appropriate module using the \textsl{Module Selector}. Next, we extract the corresponding \textsl{AugModule} and inject it into the subsequent blocks for inference. 

\section{Experiment Implementation Details}\label{sup_sec:experiment_implementation_details}
In this section, we provide additional experimental details. Firstly, to ensure reproducibility, we establish stable point-type prompts for each image, selecting positive points within the ground-truth instance segmentation mask.

Secondly, we provide details regarding the datasets used in our experiments. For medical-related tasks, we utilize the Kvasir-SEG dataset, which comprises $800$ training images and $200$ testing images, as well as the ISIC2016 dataset, containing $900$ training images and $379$ testing images. For camouflaged datasets, we employ the CAMO and COD10K datasets. The CAMO dataset has $1,000$ images, with $750$ for training and $250$ for testing. In the COD10K dataset, we focus exclusively on camouflaged data, which includes $3,040$ images and $2,026$ images for training and testing, respectively. For the shadow dataset, we use the ISTD dataset, which consists of $1,330$ training and $540$ testing images. To support reverting \method to the original SAM, we train the \textsl{Module Selector} on 100 samples from the COCO subset after CL. This subset, sourced from prior work~\cite{CLSAM}, contains 200 COCO samples. We randomly select 100 samples for training the \textsl{Module Selector} and use the remaining 100 for testing. Except the datasets we used in the main paper, we also test the performance of \method and other CL methods on CoSAM benchmark~\cite{CLSAM}.

Thirdly, since there is less prior work on CL with SAM, we integrate several existing CL methods—EWC~\cite{EWC}, ER~\cite{ER}, DER~\cite{DER}, SPPA~\cite{SPPA}, LAG~\cite{LAG}, and O-LoRA~\cite{O-Lora}—for comparison with \method. All methods train SAM and SAM2 by LoRA structures. Furthermore, we also compare \method with MoDA, a CL method for SAM. 

The implementation details for each method are as follows: (1) \textbf{EWC}: We adhere to \cite{EWC} and set the coefficient for the regulating loss to $100$. (2) \textbf{ER}: We follow original work~\cite{ER}, replaying previous samples during training with the same batch size. (3) \textbf{DER}: Based on original work~\cite{DER}, we implement DER++ and replay previous samples with the same batch size during training. (4) \textbf{SPPA}: In accordance with original work~\cite{SPPA}, we use only the alignment loss with a coefficient of $0.0001$. Other components, such as contrastive loss and pseudo labels for old classes, are not applicable in the domain-incremental CS setting, as they pertain to the class-incremental CS setting. (5) \textbf{LAG}: We fully adopt the method described in original work~\cite{LAG}, which aligns with the domain-incremental CS setting, and set all coefficients to $0.001$. (6) \textbf{O-LoRA}: We follow original work~\cite{O-Lora} and set the coefficient for the regularization loss to $0.6$. (7) \textbf{MoDA}: We strictly followed by original code~\cite{CLSAM}.

\section{Further Experiments}\label{sup_sec:further_experiments}
This section presents additional experiments not included in the main paper due to page limitations. First, we provide evidence for hyperparameters used in \method (Appendix~\ref{sup_sec:hype-parameter_selection}). Second, we present a visualized result of \method during the CL process (Appendix~\ref{sec:qualitative_results}). Thirdly, we demonstrates performance of \method across different dataset orders (Appendix~\ref{sec:different_order}). Next, we present additional evaluations on the CoSAM benchmark (Appendix~\ref{sec:cosam_datasets}), followed by an analysis of \method's ability to revert to the original SAM (Appendix~\ref{sup_sec:origin_SAM}). Lastly, we discuss the poor performance of fine-tuning methods (Appendix~\ref{sup_sec:discussion_of_comparing_methods}).

\subsection{HypeParameter Selection}\label{sup_sec:hype-parameter_selection}
\method has two main hyperparameters: the number of fine-tuning blocks and the number of stored embeddings for each learned domain.

\subsubsection{Number of Fine-tuning Blocks}\label{sec:num_blocks}
\begin{figure}[h]
    \centering
    \includegraphics[width=0.9\linewidth]{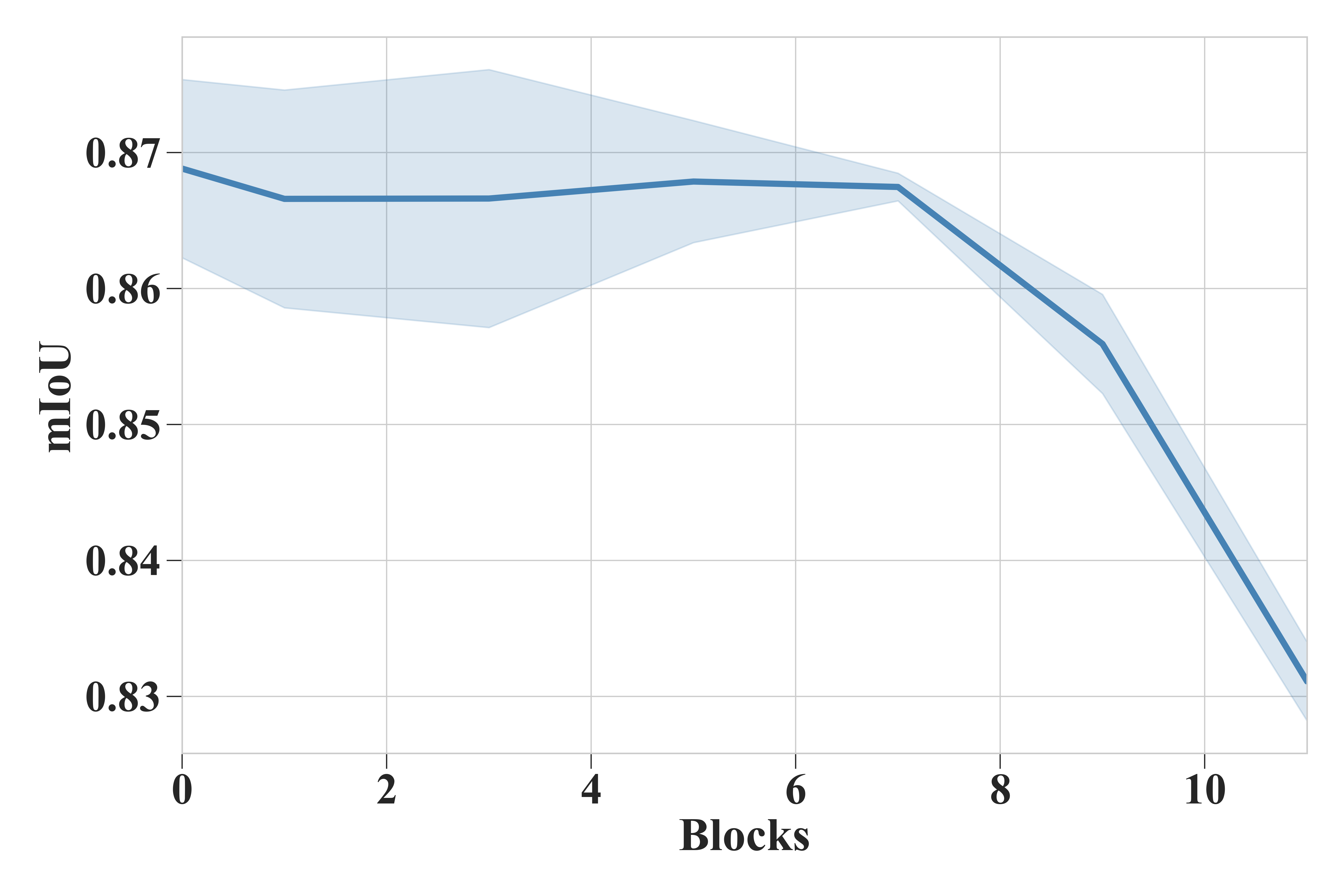}
    \caption{We tested influences of fine-tuning different numbers of last blocks across five runs on the Kvasir dataset in SAM. The x-axis indicates the starting point of fine-tuning blocks; for example, $x=1$ signifies that fine-tuning begins from the 1$^{th}$ block (inclusive). The solid line represents the average accuracy of the mean Intersection over Union (mIoU), while the error bars indicate the variation.}
    \label{fig:num_blocks}
\end{figure}

\begin{figure}[h]
    \centering
    \includegraphics[width=0.9\linewidth]{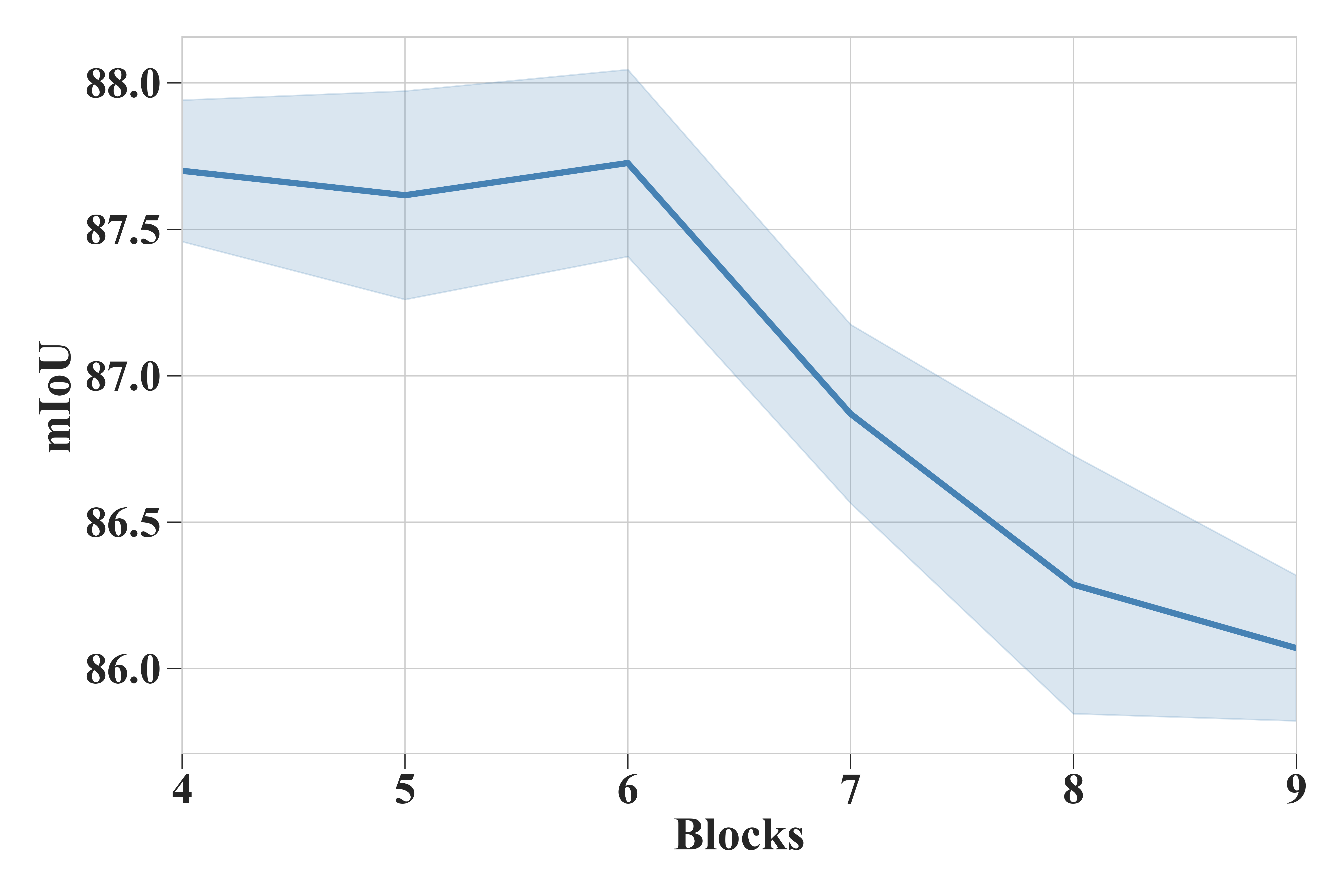}
    \caption{We tested influences of fine-tuning different numbers of last blocks across five runs on the Kvasir dataset in SAM2, similar to Figure~\ref{fig:num_blocks} of the Appendix. We only test blocks indexed from $4$ onward, as previous blocks have different dimensions that are not suitable for \textsl{SLoRA}. }
    \label{fig:num_blocks_sam2}
\end{figure}

As mentioned in the main paper, we fine-tune only the blocks following the 6$^{th}$ block in the image encoder for SAM. Meanwhile, we fine-tune only the blocks following the 5$^{th}$ block in the image encoder for SAM2. These manner are designed to reduce the latency of \textsl{Module Selector}, as \method would significantly increase latency by requiring the image encoder to be run twice: first to obtain embeddings and then to infer with \textsl{AugModule}. By fine-tuning only the subsequent blocks, we can efficiently utilize the embeddings required for \textsl{Module Selector} during continuous inference, as the added latency is limited to a simple MLP. This raises an important question: \textbf{Does fine-tuning subsequent few blocks significantly impair performance compared to fine-tuning all blocks?}

For SAM, as shown in Figure~\ref{fig:num_blocks}, the results indicate that the average accuracy of fine-tuning all blocks is similar to fine-tuning only the blocks after the 6$^{th}$ block. While fine-tuning more blocks may offer the potential for higher performance, it tends to be unstable. Therefore, considering the high accuracy in domain discrimination of middle blocks and their stability displayed in Figure~\ref{fig:num_blocks}, we opt to fine-tune only the blocks after the 6$^{th}$ block.

For SAM2, as shown in Figure~\ref{fig:num_blocks_sam2}, the results indicate that the average accuracy of fine-tuning all blocks is similar to fine-tuning only the blocks after the 5$^{th}$ block. Meanwhile, considering the high accuracy in domain discrimination of middle blocks, as noted in Figure~\ref{fig:selecting_sam2}, we opt to fine-tune only the blocks after the 5$^{th}$ block.

\subsubsection{Number of Stored Embeddings}\label{sec:discuss_stored_embeddings}

\begin{table*}[htbp]
\centering
{\adjustbox{width=\textwidth}{\scalebox{0.9}{
\begin{tabular}{c||ccc|ccc|ccc}
\hline
\multicolumn{10}{c}{Kvasir $\rightarrow$ CAMO $\rightarrow$ ISTD $\rightarrow$ ISIC $\rightarrow$ COD} \\ \hline
\multirow{2}{*}{Stored Number} & \multicolumn{3}{c|}{AA}& \multicolumn{3}{c|}{FM} & \multicolumn{3}{c}{FT} \\ 
\cline{2-10} &mIoU $\uparrow$ & mF1$\uparrow$ & mMAE$\downarrow$ & mIoU$\downarrow$ & mF1$\downarrow$ & mMAE $\uparrow$ & mIoU$\uparrow$ & mF1$\uparrow$ & mMAE $\downarrow$\\ \hline 
\multicolumn{10}{c}{SAM} \\ \hline
\textbf{10}& 0.808& 0.878& 0.037& -0.0546& -0.0557& 0.0360& 0.623& 0.735& 0.115\\
\textbf{50}& 0.820& 0.887& 0.033& 0.0072& 0.0058& -0.0016& 0.640& 0.749& 0.105\\
\textbf{100}& 0.824& 0.891& 0.032& 0.0090& 0.0077& -0.0027& 0.641& 0.749& 0.107\\
\textbf{200}& 0.831& 0.887& 0.0032& -0.00002& -0.00004& -0.0002& 0.668& 0.773& 0.089\\
\textbf{300}& 0.833& 0.898& 0.029& 0.0009& 0.0008& 0.00005& 0.656& 0.762& 0.098\\
\textbf{400}& 0.837& 0.901& 0.029& 0.0005& 0.0005& -0.0001& 0.671& 0.775& 0.088\\
\textbf{500}& 0.838& 0.902& 0.029& -0.0001& -0.0002& 0.00001& 0.670& 0.774& 0.088\\
\hline
\multicolumn{10}{c}{SAM2} \\ \hline
\textbf{10}& 0.829& 0.893& 0.036& -0.0240& -0.0186& 0.0113& 0.603& 0.714& 0.160\\
\textbf{50}& 0.833& 0.896& 0.035& -0.0364& -0.0355& 0.0260& 0.608& 0.718& 0.153\\
\textbf{100}& 0.840& 0.902& 0.030& 0.0021& 0.0018& -0.0007& 0.607& 0.717& 0.154\\
\textbf{200}& 0.845& 0.906& 0.028& -0.0012& -0.0009& 0.0006& 0.610& 0.719& 0.152\\
\textbf{300}& 0.843& 0.905& 0.028& 0.0019& 0.0017& -0.0009& 0.613& 0.722& 0.149\\
\textbf{400}& 0.844& 0.905& 0.028& 0.0014& 0.0011& 0.00008& 0.614& 0.723& 0.148\\
\textbf{500}& 0.845& 0.906& 0.027& -0.0004& -0.0002& 0.0006& 0.614& 0.723& 0.148\\
\hline
\end{tabular}}
}}
\caption{Evaluation with different numbers of stored embeddings per task.}
\label{tab:embedding_num}
\end{table*}

This section outlines the rationale for selecting the specific number of embeddings to store per domain. As shown in Table~\ref{tab:embedding_num}, we evaluated \method with $10$, $50$, $100$, $200$, $300$, $400$, and $500$ embeddings per domain using both SAM and SAM2. Notably, the performance differences across these configurations were minimal—even with as few as 10 embeddings per domain, \method still achieved strong results. This observation further validates the effectiveness of the selection process illustrated in Figure~\ref{fig:selecting} and Figure~\ref{fig:selecting_sam2}. For a balance between CL performance and storage efficiency, we selected $300$ embeddings per domain in our experiments.

\subsection{Visualized Result During CL Process}\label{sec:qualitative_results}
\begin{figure*}[htbp]
    \centering
    \includegraphics[width=0.9\linewidth]{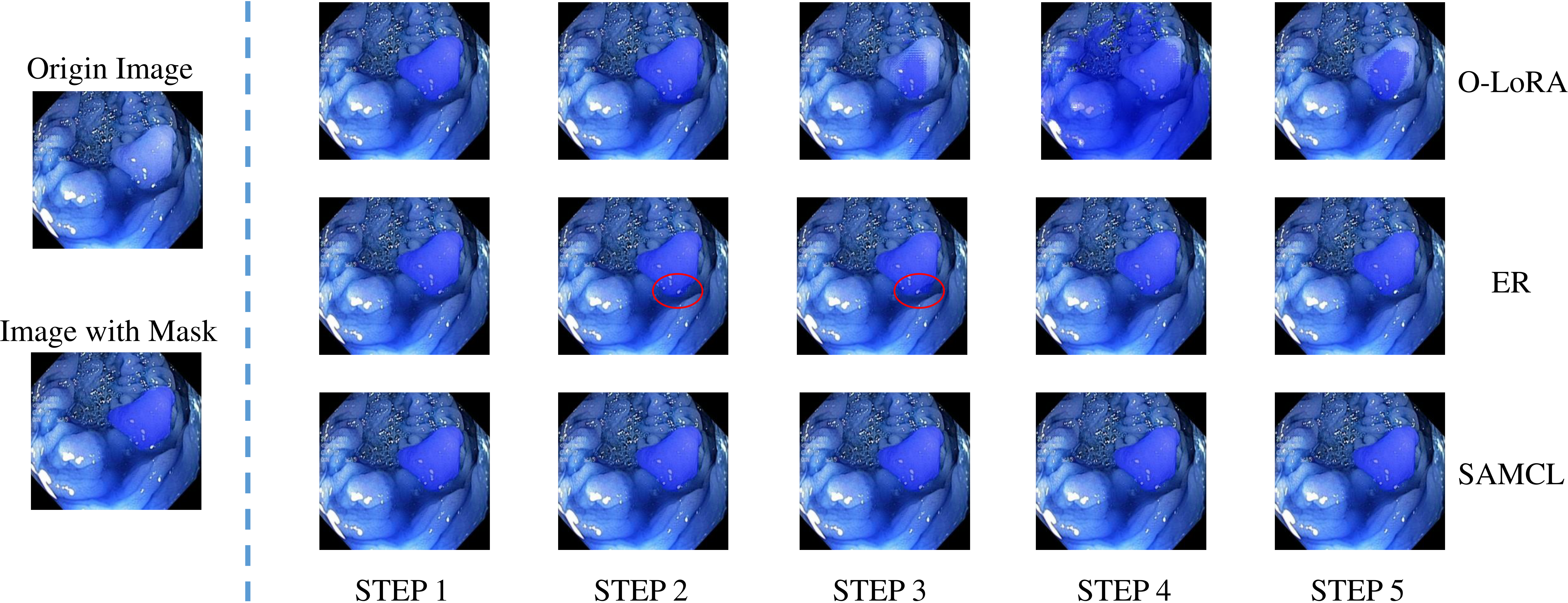}
    \caption{Qualitative evaluation during the CL process. The two figures on the left serve as references, while the five columns on the right illustrate the variation in segmentation performance for three different methods on the Kvasir dataset. \method with SAM demonstrates both consistent and excellent performance throughout the entire CL process. In contrast, O-LoRA shows a decline in performance beginning at step $3$. Although ER achieves strong performance at step $5$, it encounters issues at steps $2$ and $3$, as indicated by the red circles.}
    \label{fig:qualitive}
\end{figure*}

This section presents the visualized evaluations conducted during the CL process. The incremental datasets are ordered as Kvasir, CAMO, ISTD, ISIC, and COD, corresponding to steps $1$ through $5$. As shown in Figure~\ref{fig:qualitive}, we evaluate the Kvasir dataset throughout the entire CL process, assessing its quantitative performance after all incremental training steps. The result demonstrates the robustness of \method. Although ER demonstrates excellent performance in the final step, it still exhibits some degradation in segmentation ability for the Kvasir dataset during intermediate steps.

\subsection{Influence of Different Incremental Orders}\label{sec:different_order}

CL is challenged not only by catastrophic forgetting but also by the sensitivity to task order \cite{order}, which can lead to unstable performance.

This section evaluates the performance of \method under different task orderings. As shown in Table~\ref{tab:orders} of the Appendix, we compare the CL performance of ER, MoDA, and \method across six distinct incremental orders. The results show the excellent performance of \method in all orders, especially in average accuracy (AA) and forgetting measure (FM) metrics.
To better illustrate the robustness of each method, Figure~\ref{fig:robust} of the Appendix presents an analysis of robustness.

To provide a more comprehensive evaluation beyond continual learning (CL) metrics, we report the detailed final test results corresponding to Table~\ref{tab:main} in Table~\ref{tab:main_detail}.

\begin{figure*}[ht]
    \centering
  \begin{subfigure}{0.18\linewidth}
    \includegraphics[width=1\linewidth]{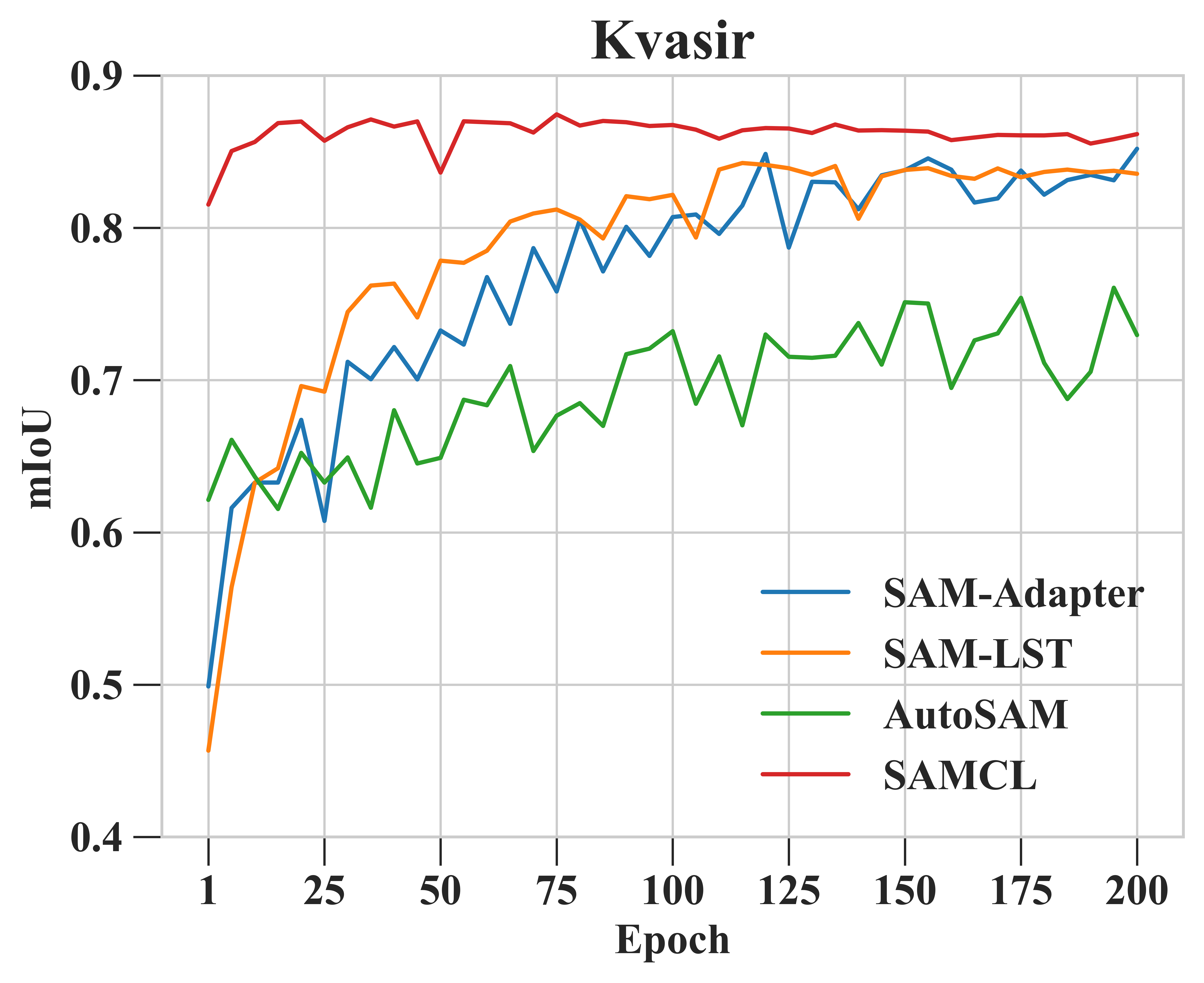}
    \caption{}
  \end{subfigure}
  \begin{subfigure}{0.18\linewidth}
    \includegraphics[width=1\linewidth]{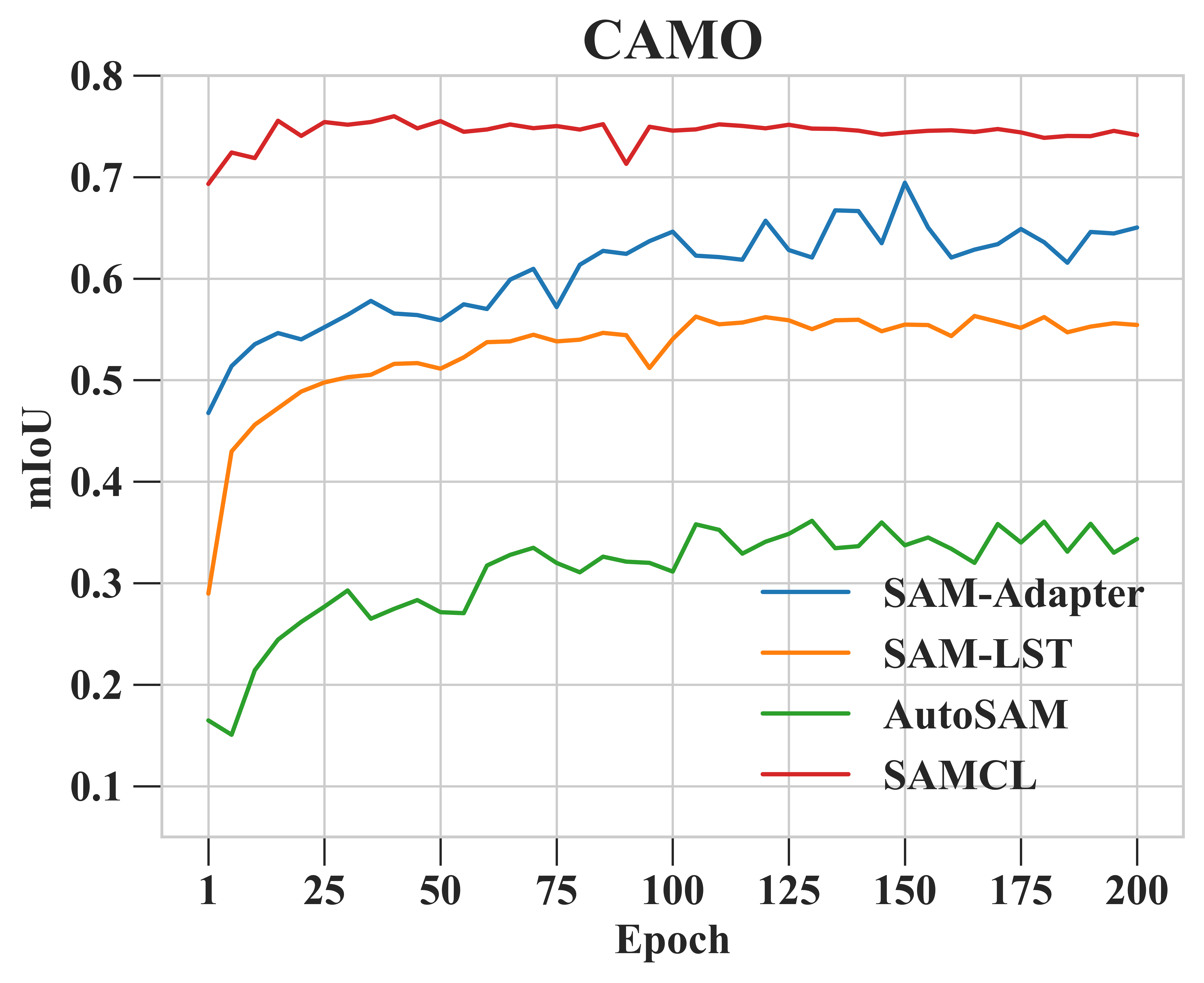}
    \caption{}
  \end{subfigure}
  \begin{subfigure}{0.18\linewidth}
    \includegraphics[width=1\linewidth]{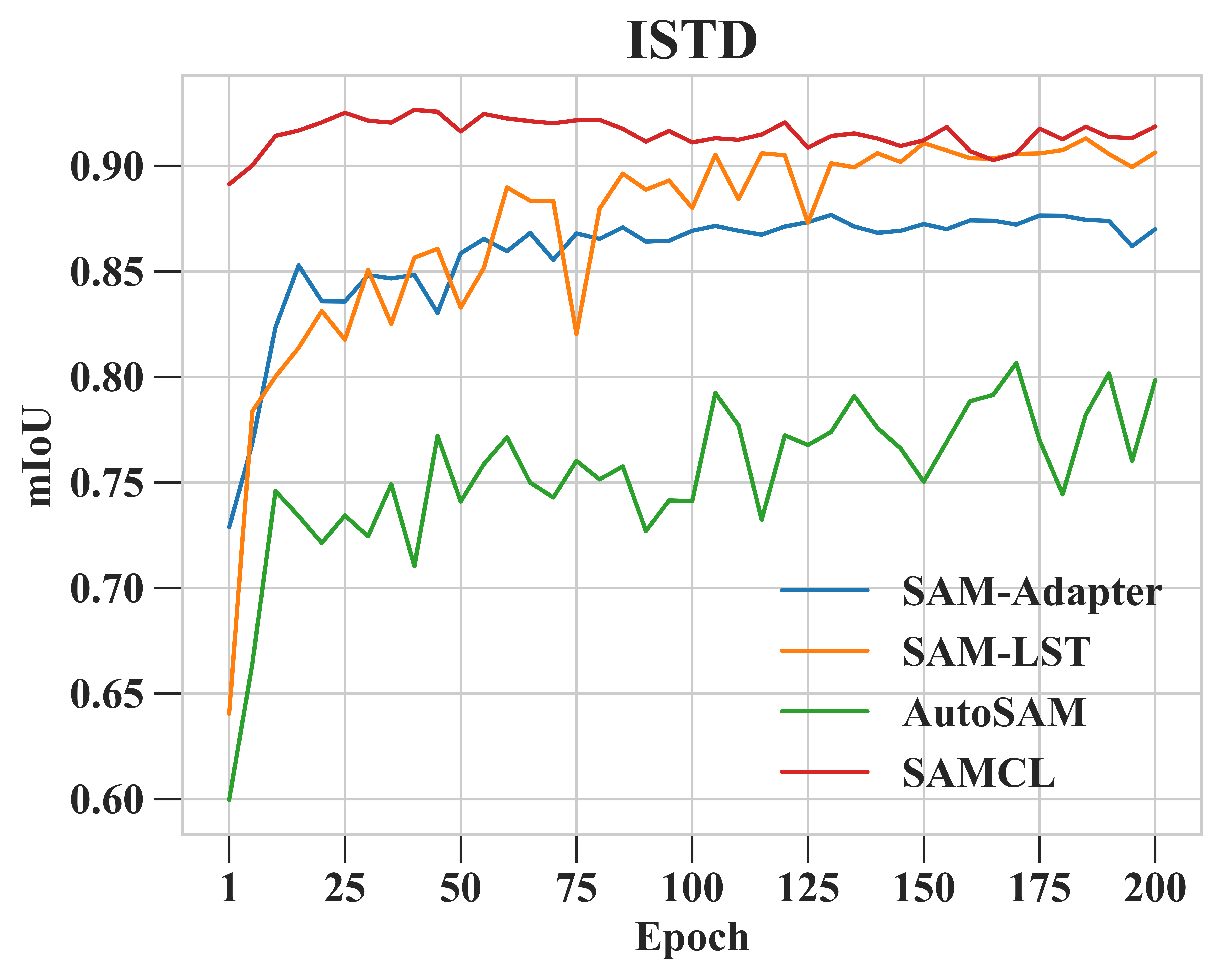}
    \caption{}
  \end{subfigure}
  \begin{subfigure}{0.18\linewidth}
    \includegraphics[width=1\linewidth]{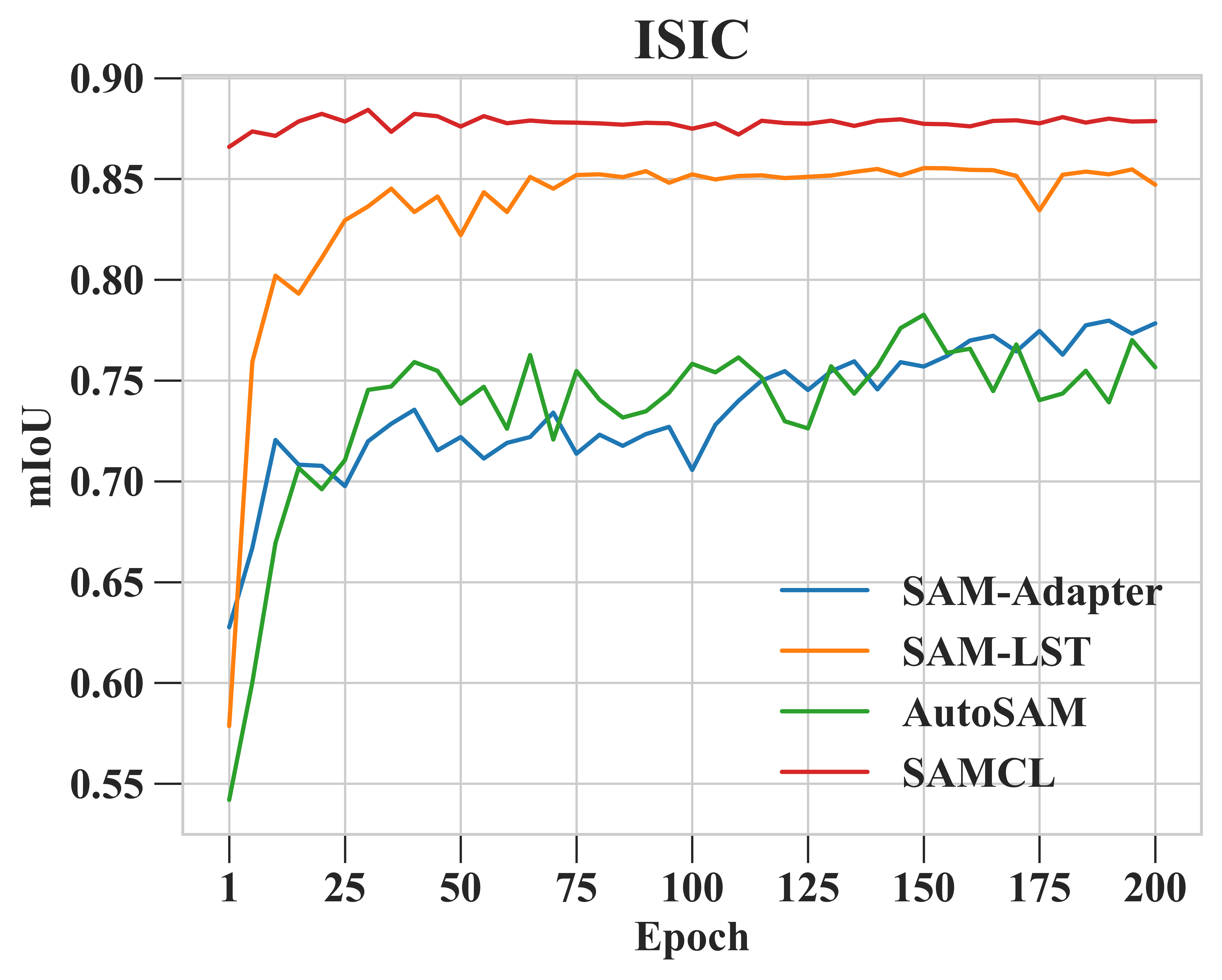}
    \caption{}
  \end{subfigure}
  \begin{subfigure}{0.18\linewidth}
    \includegraphics[width=1\linewidth]{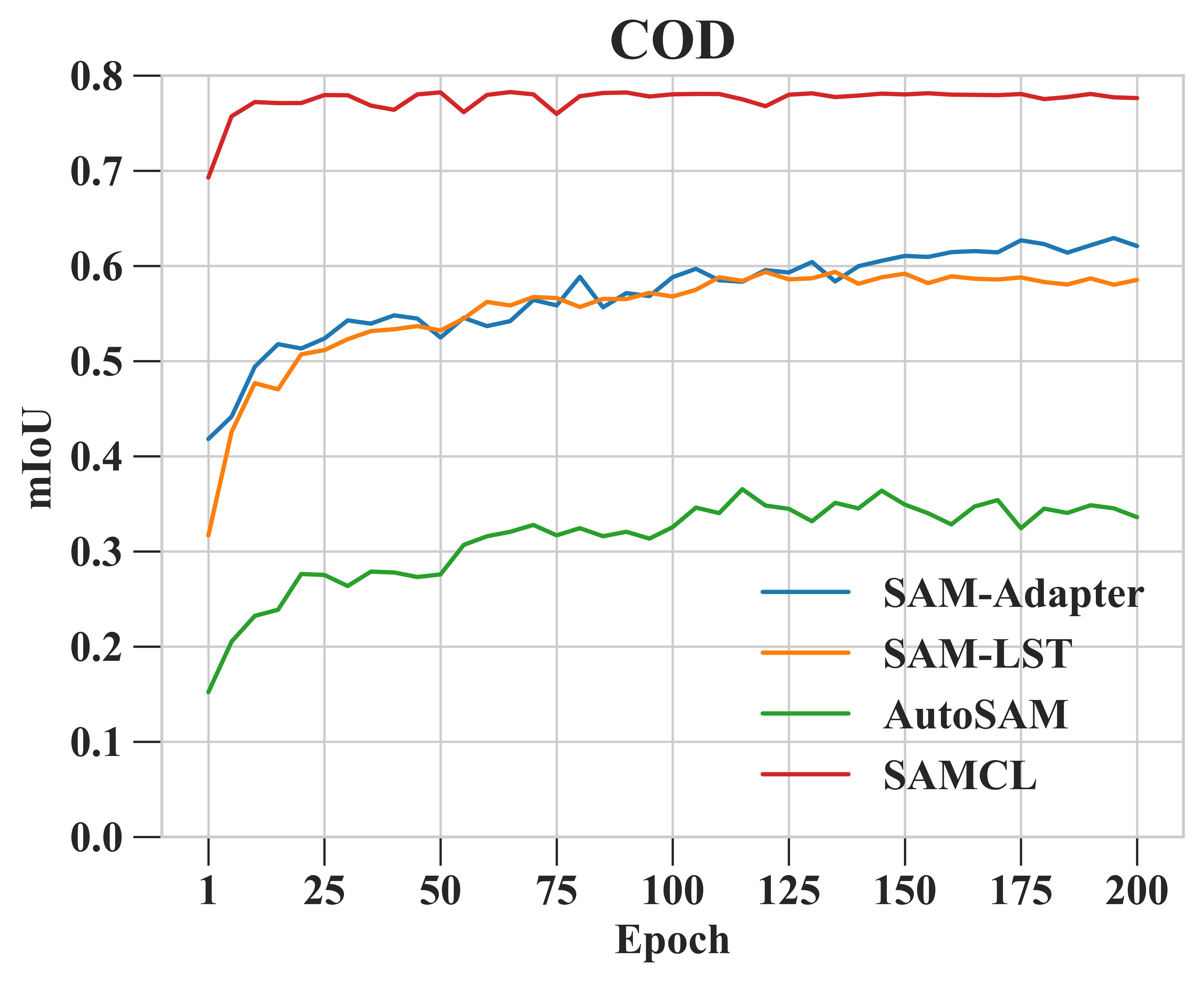}
    \caption{}
  \end{subfigure}
    \caption{Comparsion among fine-tuning methods over $200$ epochs.}
    \label{fig:abla_augmodule}
\end{figure*}

\begin{table}[ht]
\centering 
\scalebox{1}{
\begin{tabular}{c@{\hspace{2mm}}|c@{\hspace{2mm}}ccc}
\toprule 
 \textbf{Method} & \textbf{AA} $\uparrow$ & \textbf{FM} $\downarrow$ & \textbf{FT} $\uparrow$  \\
\midrule
\textbf{ER} & 0.7293 & 0.0222 & 0.5583 \\
\textbf{MoDA} & 0.7344 & 0.0141 & 0.4878 \\
\textbf{SAMCL (Ours)} & \textbf{0.7513}& \textbf{0.0018}& \textbf{0.6108}\\ 
\bottomrule
\end{tabular}
}
\caption{Evaluation results on CoSAM benchmark. The datasets order is set to BSData, Kvasir, MVTEC1, RSDD1, MVTEC2, camo, RSDD2, and MVTEC3. For simplicity, we only use mIoU metric in CL metrics AA, FM, and FT for evaluation.}
\label{tab:extra_data_main}
\end{table}

\begin{table}[ht]
\centering 
\scalebox{1}{
\begin{tabular}{c@{\hspace{2mm}}|c@{\hspace{2mm}}ccc}
\toprule 
 \textbf{Method} & \textbf{mIoU} $\uparrow$ & \textbf{mF1} $\uparrow$ & \textbf{mMAE} $\downarrow$  \\
\midrule
\textbf{SAM} & 0.6183& 0.7300& 0.0399\\
\textbf{SAMCL$_{w/o \; return}$} & 0.5281& 0.6236& 0.0590\\ 
\textbf{SAMCL} & 0.6162& 0.7272& 0.0415\\ 
\bottomrule
\end{tabular}
}
\caption{Evaluation on reverting to the original SAM. The CL setting is same to Table~1 of the main paper. As shown in the second row, applying \method without training the \textsl{Module Selector} on the COCO subset leads to a significant drop in performance compared to the original SAM. However, after training on the COCO subset, \method is able to substantially recover the original performance.}
\label{tab:return_original}
\end{table}

\begin{table*}[ht]
\centering
{\adjustbox{width=\textwidth}{\scalebox{0.75}{
\begin{tabular}{c||c@{\hspace{2mm}}c@{\hspace{2mm}}c|c@{\hspace{2mm}}c@{\hspace{2mm}}c|c@{\hspace{2mm}}c@{\hspace{2mm}}c|c@{\hspace{2mm}}c@{\hspace{2mm}}c|c@{\hspace{2mm}}c@{\hspace{2mm}}c}
\hline
\multicolumn{16}{c}{Kvasir $\rightarrow$ CAMO $\rightarrow$ ISTD $\rightarrow$ ISIC $\rightarrow$ COD} \\ \hline
\multirow{2}{*}{Method} & \multicolumn{3}{c|}{Kvasir}& \multicolumn{3}{c|}{CAMO} & \multicolumn{3}{c|}{ISTD} & \multicolumn{3}{c|}{ISIC} & \multicolumn{3}{c}{COD} \\ 

\cline{2-16} &mIoU $\uparrow$ & mF1$\uparrow$ & MAE$\downarrow$ & mIoU$\uparrow$ & mF1$\uparrow$ & MAE $\downarrow$ & mIoU$\uparrow$ & mF1$\uparrow$ & MAE $\downarrow$ &mIoU $\uparrow$ & mF1$\uparrow$ & MAE$\downarrow$ &mIoU $\uparrow$ & mF1$\uparrow$ & MAE$\downarrow$ \\ \hline
\multicolumn{16}{c}{SAM} \\ \hline

\textbf{SAM-Adapter} & 0.430& 0.564& 0.118& 0.487& 0.640& 0.124& 0.197& 0.292& 0.157& 0.536& 0.684& 0.164& 0.489& 0.643& 0.061\\
\textbf{SAM-LST} & 0.194& 0.303& 0.517& 0.323&	0.467&	0.288& 0.560& 0.685&	0.130& 	0.826&	0.898&	0.051& 0.202& 0.314& 0.268\\

\textbf{AutoSAM} & 0.201& 0.127& 0.263& 0.346& 0.239& 0.239& 0.161& 0.098& 0.263& 0.305& 0.200& 0.354& 0.355& 0.255& 0.173\\ 

\hline
\textbf{LoRA} & 0.724& 0.818& 0.066& 0.657& 0.771& 0.076& 0.646& 0.755&	0.071& 	0.748& 0.845& 0.082& 0.739&	0.834& 0.025 \\ 
\textbf{EWC} & 0.770& 0.858& 0.042& 0.682&	0.792&	0.068&	0.630& 0.744&	0.076& 	0.748&	0.846&	0.081& 0.748&	0.840&	0.024 \\
\textbf{ER} & 0.842& 0.907&	0.027& 0.692& 0.800& 0.068&	0.899& 0.939& 0.017&	0.858&	0.921&	0.040& 0.747&	0.840&	0.024\\
\textbf{DER} & 0.839&	0.905&	0.027& 0.692&	0.801&	0.065&	0.884&	0.928&	0.019&	0.860&	0.922&	0.042& 0.746&	0.838&	0.024 \\
\textbf{SPPA} & 0.314&	0.445&	0.113& 0.273& 0.398& 0.161&	0.269&	0.391&	0.147& 	0.278&	0.404&	0.225& 0.276& 0.395& 0.099 \\
\textbf{LAG} & 0.765&	0.849&	0.055& 0.696& 0.802& 0.064&	0.612&	0.747&	0.080&	0.734&	0.834&	0.087& 0.708& 0.816& 0.028 \\
\textbf{O-LoRA} & 0.749& 0.842& 0.049& 0.624& 0.746& 0.078& 0.678& 0.779& 0.060& 0.783& 0.867& 0.077& 0.689& 0.795& 0.031\\
\textbf{MoDA} & 0.829&	0.881&	0.051&	0.689&	0.795& 	0.072& 	0.727&	0.806&	0.051&	0.839&	0.899&	0.050&	0.695&	0.038&	0.038\\
\hline
\textbf{SAMCL (Ours)} & 0.864& 0.920& 0.024& 0.747& 0.841& 0.056& 0.920& 0.951& 0.011& 0.877& 0.932& 0.033& 0.772& 0.856& 0.021\\
\hline
\multicolumn{16}{c}{SAM2} \\ \hline
\textbf{LoRA} & 0.774& 0.852& 0.043& 0.654&	0.762&	0.074&	0.549&	0.674&	0.086&	0.592&	0.705&	0.144&	0.144& 0.845& 0.024 \\ 
\textbf{EWC} & 0.770& 0.855& 0.047& 0.698& 0.799& 0.067& 0.582&	0.700&	0.086&	0.675&	0.779&	0.109&	0.761& 0.850& 0.022\\
\textbf{ER} & 0.831& 0.902& 0.027&	0.684&	0.790&	0.070&	0.884&	0.928&	0.020&	0.853&	0.917&	0.043&	0.751&	0.842& 0.023\\
\textbf{LAG} &  0.759&	0.847&	0.043&	0.623&	0.732&	0.081&	0.455&	0.587&	0.101&	0.656&	0.765&	0.117&	0.737&	0.832&	0.025 \\
\textbf{O-LoRA} & 0.659& 0.774&	0.065&	0.600&	0.721&	0.087&	0.607&	0.728&	0.076&	0.716&	0.813&	0.105&	0.692&	0.800&	0.030 \\
\hline
\textbf{SAMCL (Ours)} & 0.866& 0.921& 0.026& 0.763& 0.853& 0.050& 0.935& 0.962& 0.009& 0.877& 0.931& 0.036& 0.774& 0.858& 0.021\\
\hline
\end{tabular}}}}
\caption{Final results after CL following the order of Kvasir, CAMO, ISTD, ISIC, and COD in Table~1.}
\label{tab:main_detail}
\end{table*}

\begin{table*}[ht]
\centering
{\adjustbox{width=\textwidth}{\scalebox{0.9}{
\begin{tabular}{c||ccc|ccc|ccc}
\hline
\multirow{2}{*}{Method} & \multicolumn{3}{c|}{AA}& \multicolumn{3}{c|}{FM} & \multicolumn{3}{c}{FT} \\ 
\cline{2-10} &mIoU $\uparrow$ & mF1$\uparrow$ & mMAE$\downarrow$ & mIoU$\downarrow$ & mF1$\downarrow$ & mMAE $\uparrow$ & mIoU$\uparrow$ & mF1$\uparrow$ & mMAE $\downarrow$\\ \hline 

\multicolumn{10}{c}{Kvasir $\rightarrow$ CAMO $\rightarrow$ ISTD $\rightarrow$ ISIC $\rightarrow$ COD} \\ \hline
\textbf{ER} &0.8080& 0.8818& 0.0355& 0.0107& 0.0072& -0.0034& 0.6305& 0.7481&\textbf{0.0871}\\
\textbf{MoDA} & 0.7561&	0.8357&	0.0521&	0.0208&	0.0141&	-0.0026&	0.6375&	0.7577&	0.0942\\
\textbf{SAMCL} & \textbf{0.8364}& \textbf{0.9005}& \textbf{0.0293}& \textbf{0.0019}& \textbf{0.0016}& \textbf{-0.0003}& \textbf{0.6689}& \textbf{0.7737}& 0.0894\\
\hline
\multicolumn{10}{c}{ISIC $\rightarrow$ COD $\rightarrow$ ISTD $\rightarrow$ Kvasir $\rightarrow$ CAMO} \\ \hline
\textbf{ER} &0.8026&	0.8784&	0.0360&	0.0102&	0.0070&	-0.0025& 0.5555& 0.6743& 0.1228\\
\textbf{MoDA} & 0.7678& 0.8441& 0.0496& 0.0184& 0.0155& -0.0041& \textbf{0.6647}& \textbf{0.7624}& \textbf{0.0807}\\
\textbf{SAMCL} & \textbf{0.8380}& \textbf{0.9021}& \textbf{0.0283}& \textbf{0.0009}& \textbf{0.0006}& \textbf{0.00006}& 0.6284& 0.7306& 0.0940\\ \hline

\multicolumn{10}{c}{ISIC $\rightarrow$ CAMO $\rightarrow$ COD $\rightarrow$ Kvasir $\rightarrow$ ISTD} \\ \hline
\textbf{ER} &0.8074&	0.8807&	0.0348&	0.0111&	0.0076&	-0.0019& \textbf{0.6546}& \textbf{0.7616}& \textbf{0.0907}\\
\textbf{MoDA} & 0.7570& 0.8345& 0.0530& 0.0167& 0.0145& -0.0045& 0.6460& 0.7406& 0.1163
\\
\textbf{SAMCL} & \textbf{0.8295}& \textbf{0.8957}& \textbf{0.0302}& \textbf{0.0017}& \textbf{0.0014}& \textbf{0.0004}& 0.6406& 0.7446& 0.1143\\ \hline

\multicolumn{10}{c}{Kvasir $\rightarrow$ ISIC $\rightarrow$ CAMO $\rightarrow$ COD $\rightarrow$ ISTD} \\ \hline
\textbf{ER} &0.7890&	0.8680&	0.0417&	0.0289&	0.0198&	-0.0095& 0.6364& 0.7511& 0.0900\\
\textbf{MoDA} & 0.7636& 0.8408& 0.0517& 0.0080& 0.0078& -0.0032& \textbf{0.6909}& \textbf{0.7865}& \textbf{0.0808}\\
\textbf{SAMCL} & \textbf{0.8372}& \textbf{0.9012}& \textbf{0.0284}& \textbf{0.0007}& \textbf{0.0006}& \textbf{-0.0002}& 0.6468& 0.7557& 0.1043\\ \hline

\multicolumn{10}{c}{COD $\rightarrow$ Kvasir $\rightarrow$ ISTD $\rightarrow$ CAMO $\rightarrow$ ISIC} \\ \hline
\textbf{ER} &0.8163& 0.8874& 0.0323& 0.0033& 0.0016& \textbf{0.0002}& 0.6822& 0.7895& 0.0801\\
\textbf{MoDA} & 0.7592& 0.8354& 0.0516& 0.0077& 0.0072& -0.0024& 0.6424& 0.7358& 0.0831\\
\textbf{SAMCL} & \textbf{0.8366}& \textbf{0.9005}& \textbf{0.0296}& \textbf{0.0018}& \textbf{0.0016}& -0.0002& \textbf{0.6973}& \textbf{0.7953}& \textbf{0.0797}\\ \hline

\multicolumn{10}{c}{COD $\rightarrow$ CAMO $\rightarrow$ ISIC $\rightarrow$ Kvasir $\rightarrow$ ISTD} \\ \hline
\textbf{ER} &0.8073&	0.8806&	0.0364&	0.0180&	0.0125&	-0.0053& 0.6999& 0.8013& 0.0750\\
\textbf{MoDA} & 0.7576& 0.8353& 0.0536& 0.0178& 0.0157& -0.0056& \textbf{0.7411}& \textbf{0.8247}& \textbf{0.0622}\\
\textbf{SAMCL} & \textbf{0.8290}& \textbf{0.8953}& \textbf{0.0312}& \textbf{0.0032}& \textbf{0.0028}& \textbf{-0.0005}& 0.6798& 0.7800& 0.0956\\ \hline

\end{tabular}}}}
\caption{Evaluation metrics AA, FM, and FT in different orders as depicted in Figure~\ref{fig:robust}.}
\label{tab:orders}
\end{table*}

\subsection{Evaluation on CoSAM benchmark}\label{sec:cosam_datasets}
To provide a more comprehensive evaluation, we also assess \method on the CoSAM benchmark~\cite{CLSAM}, which is designed to evaluate CL performance throughout the CL process. We compare \method against two representative methods, ER and MoDA. As shown in Table~\ref{tab:extra_data_main}, \method consistently outperforms the baselines, achieving the best overall performance.

\subsection{Return to Original SAM}\label{sup_sec:origin_SAM}
Although \method may not initially restore the original SAM perfectly, this limitation can be addressed by training the \textsl{Module Selector} on a small COCO subset to emulate a virtual module corresponding to the original SAM. In other words, the COCO subset can be treated as the first task in the continual learning process.

To verify this approach, we conduct an experiment to evaluate whether \method can effectively revert to the original SAM. As shown in Table~\ref{tab:return_original}, we test \method under different settings using the COCO subset. This subset is derived from previous work~\cite{CLSAM}, consisting of 200 COCO samples. We randomly select 100 samples to train the \textsl{Module Selector}, and use the remaining 100 to evaluate performance on this subset. The results in Table~\ref{tab:return_original} demonstrate that, after training on the COCO subset, \method successfully recovers performance nearly identical to the original SAM, with only minimal differences. These findings confirm the effectiveness of \method in reverting to the original SAM.

\subsection{Discussion of Fine-tuning Methods}\label{sup_sec:discussion_of_comparing_methods}

The poor performance of SAM-Adapter, SAM-LST, and AutoSAM, given that observed in the Table~\ref{tab:abla_augmodule}, may stem from insufficient training, as $20$ epochs are significantly fewer than what is specified in their original papers. To address this, we conducted further experiments over $200$ epochs. As shown in Figure~\ref{fig:abla_augmodule}, \textsl{AugModule} quickly reaches convergence in approximately $20$ epochs, while other methods require more than $20$ epochs to converge. Notably, although these methods take longer to converge, most still underperform compared to \textsl{AugModule}. Therefore, \textsl{AugModule} is both effective and efficient. Admittedly, the rapid convergence of \textsl{AugModule} can be attributed to the LoRA technique. However, its overall effectiveness in learning and efficiency in storage remain superior, as demonstrated in Table~\ref{tab:abla_augmodule}.

\section{Limitations and Future Works}\label{sup_sec:limitations_and_future_works}
No solution is without limitations. Despite \method achieves exremely low storage consumption during CL, incremental storage consumption still remains a challenge throughout the infinite CL process. Future research may focus on minimizing this consumption or even eliminating the need for historical information. Additionally, enhancing SAM2 for CL in video segmentation presents another promising area for exploration.

\end{document}